\documentclass[10pt,journal,compsoc]{IEEEtran}
\usepackage{graphicx}
\usepackage{comment}
\usepackage{amsmath,amssymb} 
\usepackage{color}
\usepackage{subfigure}
\usepackage{verbatim}
\usepackage{bm}
\usepackage{multirow}
\usepackage{tabulary}
\usepackage{booktabs}
\usepackage{soul}

\ifCLASSOPTIONcompsoc
  \usepackage[nocompress]{cite}
\else
  \usepackage{cite}
\fi

\begin{document}
\hyphenpenalty=5000
\tolerance=1000

\title{Beyond 3DMM: Learning to Capture High-fidelity 3D Face Shape}

\author{Xiangyu~Zhu,~\IEEEmembership{Senior Member,~IEEE,}
        Chang~Yu,
        Di Huang,~\IEEEmembership{Senior Member,~IEEE,}\\
        Zhen Lei,~\IEEEmembership{Senior Member,~IEEE,}
        Hao Wang,
        and~Stan Z. Li,~\IEEEmembership{Fellow,~IEEE}
\IEEEcompsocitemizethanks{

\IEEEcompsocthanksitem Xiangyu Zhu, Chang Yu, and Hao Wang are with Center for Biometrics and Security Research \& National Laboratory of Pattern Recognition, Institute of Automation,
Chinese Academy of Sciences, 95 Zhongguancun Donglu, Beijing 100190, China and also with the School of Artificial Intelligence, University of Chinese Academy of Sciences (UCAS), Beijing 100049, China. \protect Email: \{xiangyu.zhu, chang.yu\}@nlpr.ia.ac.cn, haowang7308@gmail.com.

\IEEEcompsocthanksitem Di Huang is with Beijing Advanced Innovation Center for BDBC, Beihang University, Beijing, China. \protect Email: dhuang@buaa.edu.cn.

\IEEEcompsocthanksitem Zhen Lei is with Center for Biometrics and Security Research \& National Laboratory of Pattern Recognition, Institute of Automation,
Chinese Academy of Sciences, 95 Zhongguancun Donglu, Beijing 100190, China, and also with the School of Artificial Intelligence, University of Chinese Academy of Sciences (UCAS), Beijing 100049, China, and also with the Centre for Artificial Intelligence and Robotics, Hong Kong Institute of Science \& Innovation, Chinese Academy of Sciences, Hong Kong, China.  \protect Email: zlei@nlpr.ia.ac.cn.

\IEEEcompsocthanksitem Stan Z. Li is with the School of Engineering, Westlake University, Hangzhou, China. \protect Email: Stan.ZQ.Li@westlake.edu.cn.
}

\thanks{Manuscript received 1 Apr. 2021; revised 6 Jan. 2022; accepted 28 Mar. 2022.}
\thanks{(Corresponding author: Zhen Lei.)}
\thanks{Recommended for acceptance by D. Samaras.}
\thanks{Digital Object Identifier no. 10.1109/TPAMI.2022.3164131}
}

\markboth{IEEE TRANSACTIONS ON PATTERN ANALYSIS AND MACHINE INTELLIGENCE,~Early Access,~ 2022}
{Shell \MakeLowercase{\textit{Zhu et al.}}: Beyond 3DMM: Learning to Capture High-fidelity 3D Face Shape}

\IEEEtitleabstractindextext{%
\begin{abstract}
3D Morphable Model (3DMM) fitting has widely benefited face analysis due to its strong 3D priori. However, previous reconstructed 3D faces suffer from degraded visual verisimilitude due to the loss of fine-grained geometry, which is attributed to insufficient ground-truth 3D shapes, unreliable training strategies and limited representation power of 3DMM. To alleviate this issue, this paper proposes a complete solution to capture the personalized shape so that the reconstructed shape looks identical to the corresponding person. Specifically, given a 2D image as the input, we virtually render the image in several calibrated views to normalize pose variations while preserving the original image geometry. A many-to-one hourglass network serves as the encode-decoder to fuse multiview features and generate vertex displacements as the fine-grained geometry. Besides, the neural network is trained by directly optimizing the visual effect, where two 3D shapes are compared by measuring the similarity between the multiview images rendered from the shapes. Finally, we propose to generate the ground-truth 3D shapes by registering RGB-D images followed by pose and shape augmentation, providing sufficient data for network training. Experiments on several challenging protocols demonstrate the superior reconstruction accuracy of our proposal on the face shape.
\end{abstract}

}

\maketitle

\IEEEdisplaynontitleabstractindextext

%
%

\IEEEraisesectionheading{\section{Introduction}\label{sec:introduction}}
The core problem addressed in this paper is high-fidelity 3D face reconstruction from a single face image. As a detailed and interpretable description of face images, 3D faces have acted as an essential priori in many face-related tasks, e.g., face recognition~\cite{echeagaray2017method,liu2020joint}, face manipulation~\cite{shu2017neural,geng20193d}, expression analysis~\cite{fang20123d,bejaoui2017fully} and facial animation~\cite{karras2017audio,cudeiro2019capture}. Although there have been significant advances in 3D face reconstruction~\cite{16_feng2018joint,zhu2019face,15_Tran_2018_CVPR,guo2020towards}, the limited performance still suffers from insufficient ground-truth 3D face data. There are three processes required in ideal 3D face collection: 1) Instantly capturing the multiview 3D scans of one face. 2) Fusing the scans to a full 3D face with annotated landmarks.  3) Acquiring the topology-uniform 3D mesh by registering the full 3D face to a template. The requirements of expensive devices, fully controlled environment and laborious human labeling limit the large-scale collection of 3D faces for neural network training.

\begin{figure}
	\begin{center}
    \includegraphics[width=0.45\textwidth]{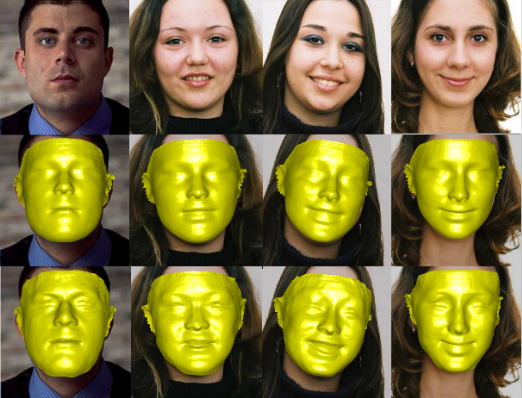}
    \end{center}
\caption{The first, second and third rows show the images, the widely applied 3DMM fitting~\cite{16_feng2018joint} results, and our results, respectively.}
\label{top}\label{fig-demo-intro}
\end{figure}

Generally, there are two strategies to tackle the data challenge: one is the 2D-to-3D method and the other is the self-supervised method. The 2D-to-3D method constructs 3D faces by fitting a 3D Morphable Model (3DMM) to annotated landmarks~\cite{zhu2019face}. The generated 3D faces overlap the face region well with pixel-level parsing accuracy. However, as a linear model built from insufficient data, 3DMM spans restricted demographic features. In previous decades, almost all the models~\cite{Paysan-AVSS-09,Alpha3_cao2014facewarehouse,yang2020facescape} cover no more than $1,000$ subjects, which is far from adequate to approximate the large diversity of human faces. Besides, in 3DMM fitting, only facial features are coarsely constrained by the landmarks, and the other large regions, such as cheek, are less constrained by shape priori. Therefore, restricted by the coarse 3D ground truth, the trained neural networks~\cite{16_feng2018joint,zhu2019face,Jackson2017Large} cannot capture personalized shape and suffer from model-like reconstruction results, as shown in Fig.~\ref{fig-demo-intro}.

On the other hand, following the analysis-by-synthesis manner, the self-supervised method~\cite{15_Tran_2018_CVPR,tran2019learning,zhou2019dense} learns a non-linear 3D face model from a large set of unlabeled images. Specifically, given a face image as input, a network encoder estimates the projection, lighting, shape, and albedo parameters, and two decoders serve as the non-linear 3DMM to map the parameters to a 3D face. Then, an analytically-differentiable rendering layer is employed to minimize the difference between the input face and the reconstructed face, and the learned decoder becomes the 3D face model. In this manner, the non-linear model can cover a larger shape space than the linear 3DMM. However, the improvement is still not satisfactory without straightforward supervision signals from ground-truth 3D faces. First, the successful reconstruction of the original face does not promise that the underlying 3D shape is the ground truth due to 2D-to-3D ambiguity. Second, the analysis-by-synthesis method mainly optimizes pixel values. Thus, the person-specific geometry, which does not contribute much to pixel values, is not well captured during optimization.

Beyond 2D-to-3D and self-supervised strategies, there is little research on how to enable neural networks to capture fine-trained geometry to improve visual verisimilitude. This paper explores this nontrivial problem and proposes that fully supervised learning is still a promising solution given only thousands of single-view RGB-D images. To this end, we extend the recent state-of-the-art methods and make the following contributions:

\textbf{Data:} We explore the construction of large-scale high-fidelity 3D face data with single-view RGB-D face images. Although it is expensive to acquire high-precision 3D faces, single-view RGB-D images can be considered an appropriate alternative because they are much easier to collect, especially considering the development of hand-held depth cameras (such as iPhone X), which strongly increases the possibility of the massive collection of medium-precision 3D data. Therefore, we register the RGB-D images and investigate two data augmentation strategies, view simulation and shape transformation, to generate a large 3D dataset \textbf{F}ine-\textbf{G}rained \textbf{3D} face (\textbf{FG3D}).

\textbf{Network:} A \textbf{F}ine-\textbf{G}rained reconstruction \textbf{Net}work (\textbf{FGNet}) is proposed in specific. With an initial 3DMM fitting result, we render the input image in $5$ calibrated views and fuse the encoded features by a many-to-one hourglass network, where the mid-level features are aligned to a common UV space to ensure identical topology among the fused features. With lossless pose normalization and receptive field alignment, the network can concentrate on shape information and encode sophisticated features to capture the personalized shape.

\textbf{Loss Function:} Although commonly effective in 3D face alignment~\cite{zhu2019face, 16_feng2018joint, cheng2020faster}, the Mean Squared Error (MSE) loss is insensitive to fine-grained geometry, which contributes significantly to the visual effect but little to shape morphing. Considerable works~\cite{tian2019contrastive,tatarchenko2019single,jin2020dr,xu20203d} also indicate that MSE cannot model how humans observe a 3D object. To address this issue, we propose a plaster sculpture descriptor to model the visual effect, where the reconstructed 3D face is rendered into several views with shading as the texture, leading to prominent improvements in visual verisimilitude.

A preliminary version of this work was published in~\cite{zhu2020beyond}. We extend it in the following aspects: 1) For the network structure, we propose a brand-new Virtual Multiview Network (VMN) with a virtual multiview camera system and a many-to-one hourglass network, which remedies the defects of the camera view network and the model view network in~\cite{zhu2020beyond}. 2) For the loss function, we improve the MSE loss by modeling the visual effect through a novel plaster sculpture descriptor. 3) For the data construction, we augment shape variations by transforming the underlying 3D shapes of face images and refining the face appearances accordingly, which further strengthens shape robustness. 4) Additional experiments are conducted to better analyze the motivation behind the design of the network structure and the loss function.

\section{Related Work}
There has been significant progress in 3D face reconstruction in recent decades. This section briefly reviews previous studies related to our proposal, including 3D morphable model fitting, non-linear 3D model construction, and fine-grained geometry reconstruction.

\subsection{3D Morphable Model Fitting}
In early years, some methods use CNN to directly estimate 3D Morphable Model parameters~\cite{zhu2019face,richardson2017learning,guo2020towards} or its variants~\cite{bas20173d,dou2017end,tewari2017mofa,tuan2017regressing,Bhagavatula2017Faster}, providing both dense face alignment and 3D face reconstruction results. However, the performances of these methods are restricted by the limitation of linear 3D space~\cite{dou2017end,hassner2013viewing,kemelmacher20113d,Alpha4_2017dense,sela2017unrestricted,S20163D}. It is also challenging to estimate the required face transformations, including perspective projection and 3D thin plate spline transformation. Recently, vertex regression networks~\cite{Jackson2017Large,16_feng2018joint}, which bypass the limitation of the linear model, have achieved state-of-the-art performance on their respective tasks. VRN~\cite{Jackson2017Large} proposes to regress 3D faces stored in the volume, but the redundant volumetric representation loses the semantic meanings of 3D vertices. PRNet~\cite{16_feng2018joint} designs a UV position map, a 2D image recording the 3D facial point cloud while maintaining the semantic meaning at each UV position. Cheng et al.~\cite{cheng2020faster} encode an image with a CNN and decode the 3D geometry with a lightweight Graph Convolutional Networks (GCN) for efficiency. Although breaking through the limitations of 3DMM, their reconstruction results are still model-like as their ground truth still comes from 3DMM fitting, e.g., 300W-LP~\cite{zhu2019face}.

\subsection{Non-linear 3D Face Model by Self-supervision}
Another way to bypass the limitation of 3DMM is to encode the 3D face model by a neural network, which can be learned on the unlabeled images in an analysis-by-synthesis manner. Tran et al.~\cite{15_Tran_2018_CVPR} achieve a certain breakthrough by learning a nonlinear model from unlabeled images in a self-supervised manner. They first estimate the projection, shape, and texture parameters with an encoder and then adopt two CNN decoders as the nonlinear 3DMM to map the parameters to a 3D textured face. With an analytically-differentiable rendering layer, the network can be trained in a self-supervised way by comparing the reconstructed images with the original images. Zhou et al.~\cite{zhou2019dense} follow a similar path but decode the shape and texture directly on the mesh with graph convolution for efficiency. Gao et al.~\cite{gao2020semi} utilize a CNN encoder and a GCN decoder to learn 3D faces from unconstrained photos with photometric and adversarial losses. In addition to learning a single model accounting for all shape variations, some works construct facial details independently. Chen et al.~\cite{chen2020self} un-warp image pixels and facial texture to the UV plane with a fitted 3DMM and predict a displacement map of the coarse shape by minimizing the difference between the reconstructed and the original images. Wang et al.~\cite{wang2020low} adopt a similar method but predict the moving rate in the direction of the vertex normal instead of the displacement map. The self-supervised nonlinear 3DMMs can cover a larger shape space than linear 3DMMs, but the fine-grained geometry, as a subtle constituting factor of face appearance, is easy to ignored due to the absence of straightforward training signals from the ground-truth shapes.

\subsection{Fine-grained Geometry Reconstruction}
Considering the lack of high-precision 3D scans, a common strategy for fine-grained reconstruction is \textbf{S}hape \textbf{f}rom \textbf{S}hading (\textbf{SfS}), which recovers 3D shapes from shading variations in 2D images. However, traditional SfS methods largely depend on the prior geometric knowledge and suffer from the ambiguity problem caused by albedo and lighting estimation. To address this issue, Richardson et al.~\cite{richardson2017learning} refine the depth map rendered by a fitted 3DMM with the SfS criterion to capture details, where a trained depth refinement net directly outputs high-quality depth maps without calculating albedo and lighting information. DF$^2$Net~\cite{zeng2019df2net} further introduces depth supervision from RGB-D images to regularize SfS in a cascaded way. However, these methods only learn the $2.5$D depth map and lack correspondence with the 3D face model. Jiang et al.~\cite{jiang20183d} propose to refine the coarse shape with photometric consistency constraints to generate a more accurate initial shape. Li et al.~\cite{li2018feature} employ a masked albedo prior to improve albedo estimation and incorporate the ambient occlusion behavior caused by wrinkle crevices. Chen et al. ~\cite{chen2019photo} construct a PCA model for details using collected scans and adopt the analysis-by-synthesis strategy to refine the PCA model. Guo et al.~\cite{guo2018cnn} employ a FineNet to reconstruct the face details, whose ground truth is synthesized by inverse rendering. Although the methods capture fine-level geometric details (such as wrinkles), their global shapes still come from the 3DMM fitting. Unlike SfS, our method aims to improve the global visual effect such as facial feature topology and face contour.

In addition to single-view reconstruction, high-quality 3D faces can be better recovered with multiview inputs. Many 3D face datasets, e.g., D3DFACS~\cite{cosker2011facs} and FaceScape~\cite{yang2020facescape}, use the calibrated depth camera arrays to reconstruct high-quality 3D shapes. Besides, recent deep learning based works have shown that multiview images of the same subject benefit shape reconstruction. DECA~\cite{feng2020learning} proposes a detail-consistency loss between different images to disentangle expression-dependent wrinkles from person-specific details. MVF-Net uses the MultiPIE~\cite{gross2010multi} samples collected by $15$ cameras for multiview 3DMM parameters regression. Shang et al.~\cite{shang2020self} also adopt the MultiPIE to alleviate the pose and depth ambiguity during 3D face reconstruction. Lin et al.~\cite{lin2020high} take the RGB-D selfie videos captured by iPhone X to reconstruct high-fidelity 3D faces. Although our proposal focuses on 3D reconstruction from a single image, these multiview systems motivate our network structure to observe the input image in several calibrated views.

\section{Background of 3D Face Reconstruction}
This section describes the formulation of our main task and other related background information. In general, a 3D face can be decomposed into \emph{coarse shape}, \emph{personalized shape} and \emph{pose}. The coarse shape is usually formulated by the seminal work of the 3D Morphable Model (3DMM)~\cite{1_blanz2003face}, which describes the 3D face space with PCA:
\begin{equation}
\mathbf{S}=\mathbf{\overline{S}} + \mathbf{A}_{id}\bm{\alpha}_{id} + \mathbf{A}_{exp}\bm{\alpha}_{exp},
\end{equation}
where $\mathbf{\overline{S}}$ is the mean shape, $\mathbf{A}_{id}$ is the principle axes trained on the 3D face scans with neutral expression and $\bm{\alpha}_{id}$ is the shape parameter, $\mathbf{A}_{exp}$ is the principle axes trained on the offsets between expression scans and neutral scans and $\bm{\alpha}_{exp}$ is the expression parameter.  Although widely implemented in various tasks~\cite{cheng2020towards, hu2017avatar, wang2017real}, the coarse shape covers limited shape variations, leading to model-like reconstruction results. The personalized shape contains most person-specific shape morphing, which can be formulated as the vertex displacement $\Delta\mathbf{S}$ missed by the linear subspace. Finally, the pose, which is formulated by the camera parameters $\mathbf{C}=[f, \mathbf{R}, \mathbf{t}_{3d}]$, determines the rigid transformation to the image plane:
\begin{equation}\label{equ-project}
\mathcal{V}(\mathbf{C}, \mathbf{S}+\Delta\mathbf{S}) = f * \mathbf{R}*(\mathbf{S}+\Delta\mathbf{S}) +\mathbf{t}_{3d},
\end{equation}
where $\mathcal{V}(\cdot, \cdot)$ is the rigid transformation, $f$ is the scale factor, $\mathbf{R}$ is the rotation matrix, and $\mathbf{t}_{3d}$ is the 3D translation vector.  Among these 3D components, camera parameter $\mathbf{C}$ and coarse shape $\mathbf{S}$ can be well estimated from a single image by recent 3DMM fitting methods~\cite{zhu2019face,16_feng2018joint,guo2020towards}. However, there is little attention paid to the recovery of personalized shape $\Delta\mathbf{S}$ due to the shortage of training data and sophisticated recovery methods, which stimulates our motivation. It is worth noting that, different from the multiview reconstruction task~\cite{lin2020high, shang2020self}, which has a well-defined solution by stereo vision, single-view reconstruction is an ill-posed problem due to the loss of depth information. Thus, the goal of our task is \emph{predicting} the personalized shape by the knowledge learned from large-scale training data.

To make CNN concentrate on personalized shape, we fit an \emph{off-the-shelf} 3DMM~\cite{zhu2019face} to estimate the coarse shape $\mathbf{S}$ and the camera parameter $\mathbf{C}$. Then our task can be formulated as:
\begin{equation}\label{eqn-task}
\arg \min\limits_{\theta} \| \mathcal{F}(\mathbf{S}^{*}) - \mathcal{F}(\emph{Net}(\mathbf{I}, V(\mathbf{S}, \mathbf{C}); \theta) + \mathbf{S}) \|,
\end{equation}
where $\mathbf{S}^{*}$ is the ground-truth 3D shape, $\mathcal{F}(\cdot)$ is a 3D feature accounting for reconstruction quality, and the network $\emph{Net}(\mathbf{I}, \mathcal{V}(\mathbf{S}, \mathbf{C});\theta)$, with $\theta$ as its network weights, takes the image $\mathbf{I}$ and the 3DMM fitting result $\mathcal{V}(\mathbf{S}, \mathbf{C})$ as the input and predicts the personalized shape. It can be seen that the network structure $\emph{Net}(\cdot)$ and the supervisory signals $\mathcal{F}(\cdot)$ are two critical topics in this formulation.

\section{Virtual Multiview Network}\label{sec-network}
\begin{figure*}[htb]
 \subfigure{
  \label{fig-overview-a}}
  \subfigure{
  \label{fig-overview-b}}
  \subfigure{
  \label{fig-overview-c}}
	\begin{center}
		\includegraphics[width=0.99\textwidth]{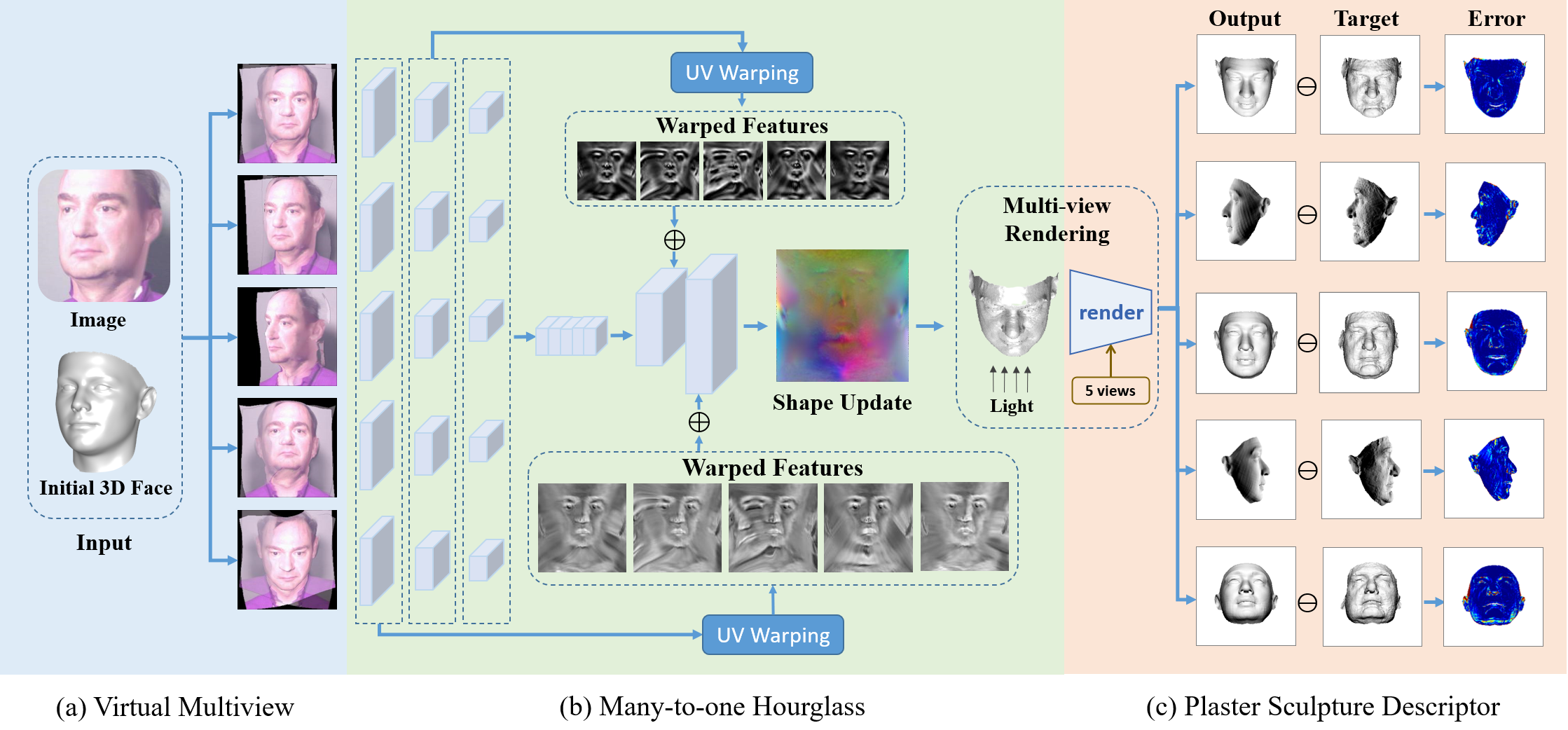}
	\end{center}
	\caption{An overview of the High-Fidelity Reconstruction Network (HRNet). (a) The input image is rendered to $5$ calibrated views to normalize pose variations while preserving the original image geometry. (b) A many-to-one hourglass network serves as the encode-decoder to fuse the multiview features and generate the vertex displacements as the personalized shape. (c) A plaster sculpture descriptor is utilized to optimize the visual effect by comparing the multiview images rendered from the shapes.}
	\label{fig-overview}
\end{figure*}

The core problem of designing the network, i.e., $\emph{Net}(\mathbf{I}, \mathbf{V})$ in Eqn.~\ref{eqn-task}, is how to fully utilize the visual information of the input image $\mathbf{I}$ and the 3D prior embedded in the fitted 3DMM $\mathbf{V}$. The network should have three properties: 1) \textbf{Normalization:} The non-shape components of the input face should be normalized with the initially fitted 3DMM to make the network concentrate on capturing shape information. 2) \textbf{Lossless:} the geometry and texture in the original image should be preserved to provide complete information for the network. 3) \textbf{Concentration:} the receptive field of each outputted vertex should cover the most related image region to highlight the subtle appearance. In this section, we motivate our \textbf{V}irtual \textbf{M}ulti-view \textbf{N}etwork (\textbf{VMN}) from the multiview camera system used to generate complete 3D scans, where a volunteer sits in a fully constrained environment and several cameras circling around the head take the pictures simultaneously. Fig.~\ref{fig-overview} shows an overview of the proposed network.

\subsection{Multiview Simulation}\label{sec-multiview-simulation}
\begin{figure}[htb]
  \subfigure{
  \label{fig-multiview-input-a}}
  \subfigure{
  \label{fig-multiview-input-b}}
  \subfigure{
  \label{fig-multiview-input-c}}
  \subfigure{
  \label{fig-multiview-input-d}}
  \subfigure{
  \label{fig-multiview-input-e}}
  \subfigure{
  \label{fig-multiview-input-f}}
	\begin{center}
		\includegraphics[width=0.49\textwidth]{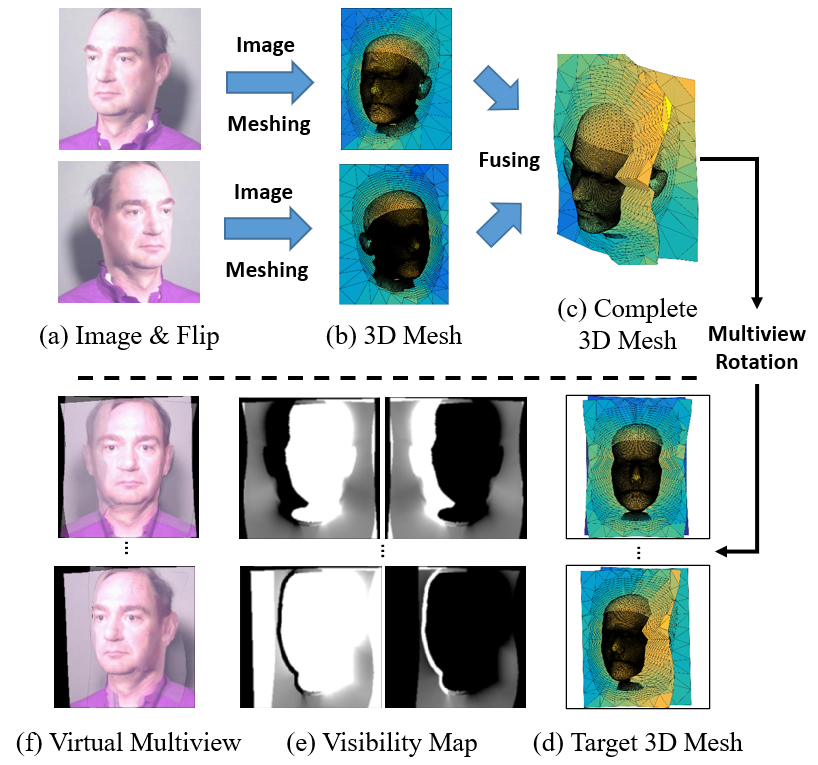}
	\end{center}
	\caption{An overview of multiview simulation. (a) The input image and the flipped one; (b) The 3D image mesh; (c) The complete 3D image mesh; (d) Rotating the 3D mesh to the target view; (e) The visibility map of the original image and the flipped image; (f) The generated virtual view. }
	\label{fig-overview-tex-icp}
\end{figure}

Given a face image, we aim to simulate its appearances at $5$ constant poses, whose $($pitch, yaw$)$ angles are $(0^{\circ},0^{\circ})$, $(0^{\circ},25^{\circ})$, $(0^{\circ},50^{\circ})$, $(15^{\circ},0^{\circ})$ and $(-25^{\circ},0^{\circ})$, as shown in Fig.~\ref{fig-overview-a}. The view synthesis can be achieved by transferring the image $\mathbf{I}$ to a 3D object through the strong prior from the fitted 3DMM $\mathbf{V}$. Following face profiling~\cite{zhu2019face}, we tile anchors on the background, set their depth to the mean of the 3D face, and triangulate them to a 3D mesh $\mathbf{V}^{I}$, which can be rendered at any required views, as shown in Fig.~\ref{fig-multiview-input-b}.
However, the texture in 3D mesh $\mathbf{V}^{I}$ is not complete. When the target pose is smaller than the original pose, the self-occluded region will be exposed, leading to large artifacts. To fill the occluded region, we also generate a flipped 3D mesh $\mathbf{V}^{I}_{flip}$ from the mirrored image and register it to $\mathbf{V}^{I}$ to complete the face texture, as shown in Fig.~\ref{fig-multiview-input-c}.

When rendering the 3D mesh at specified views, we find that each pixel is determined by either the original image or the flipped image, depending on the vertex visibility. The visibility score of each vertex is defined in terms of the angle between the vertex normal and the view direction:
\begin{equation}\label{equ-vis-score}
\emph{vis}(\mathbf{v}) =
\left\{
\begin{aligned}
&\mathbf{l}^{T}\cdot \mathcal{N}(\mathbf{v}) + 2 ~~~ if~~~ \mathbf{v} \in face\\
&\mathbf{l}^{T}\cdot \mathcal{N}(\mathbf{v}) ~~~~~~~~~  if~~~ \mathbf{v} \in background,\\
\end{aligned}
\right.
\end{equation}
where $\mathbf{l}=[0,0,1]$ is the view direction and $\mathcal{N}(\mathbf{v})$ is the normal of vertex $\mathbf{v}$ in the original image. Note that the visibility scores of face vertices are increased by $2$ to ensure that face regions overlap the background. The visibility scores, regarded as the face texture, are then rendered to the target view to obtain the visibility map, as shown in Fig.~\ref{fig-multiview-input-e}.  Finally, the virtual face image at a specified view is constructed by:
\begin{align}\label{equ-vmn-render}
\mathbf{I}_{v}=\mathbf{\lambda} \odot \mathcal{R}[\mathcal{V}(\mathbf{V}^{I}, \mathbf{C}_{v})]+\mathbf{\lambda}_{flip}\odot\mathcal{R}[\mathcal{V}(\mathbf{V}^{I}_{flip}, \mathbf{C}_{v})], ~with~ \notag\\
\begin{aligned}
&\mathbf{\lambda}(x,y)=1, \lambda_{flip}(x,y)=0 ~~~~~~ if ~ \mathbf{vis}(x,y) \ge \mathbf{vis}_{flip}(x,y)\\
&\mathbf{\lambda}(x,y)=0, \lambda_{flip}(x,y)=0.5 ~~~ if ~ \mathbf{vis}(x,y) < \mathbf{vis}_{flip}(x,y),
\end{aligned}
\end{align}
where $\mathcal{R}(\cdot)$ renders the 3D mesh $\mathcal{V}(\mathbf{V}^{I}, \mathbf{C}_{v})$ to a specified view, indicated by the camera parameter $\mathbf{C}_{v}$, $\lambda$ is the weight map calculated by comparing the visibility maps $\mathbf{vis}$, and $\odot$ is the element-wise production. As shown in Fig.~\ref{fig-multiview-input-f}, the invisible region is made transparent to indicate that the texture is not real but inpainted by face symmetry, which should guide the network to concentrate more on the visible texture. In the case that the input is the frontal view, we directly stretch the texture as the side-face texture, and the neural network is trained to handle the artifacts.

\subsection{Many-to-one Hourglass}
To fuse the features extracted from multiview inputs, we propose a novel many-to-one hourglass network with a $5$-stream encoder and a $1$-stream decoder, as shown in Fig.~\ref{fig-overview-b}. The encoder learns each view by a weight-shared CNN, and then concatenates the features as a multiview description. Afterwards, the decoder deconvolves the feature to the UV displacement map~\cite{zhu2020beyond}. Besides, the intermediate features at symmetric layers are added together to merge high-level and low-level information. However, the particular structure of the many-to-one hourglass network poses a critical challenge for intermediate feature fusion. The features at one position have different semantic meanings, since the encoder has $5$ different image views and the decoder is on the UV plane. Directly adding the features would degrade each other. To address the problem, each feature map in the encoders is warped to the UV plane according to the fitted 3DMM before being sent to the decoder.

Regarding the three properties desired for network design, the VMN fulfills the normalization property by transferring the input face to constant views. It also possesses the lossless property as little information is lost during the construction of virtual multiview input, i.e., the original topology and the external face regions are preserved. For the concentration property, since the intermediate features are all aligned to the UV plane, the receptive field can consistently concentrate on the most related region.

\section{Loss Function}\label{sec-loss}
The loss function directly judges the reconstruction results according to the targets. Generally, the Mean Square Error (MSE) loss is employed to diminish the 3D coordinate error. Given the ground-truth shape $\mathbf{S}^{*}$, the initial 3D shape $\mathbf{S}$ and the output shape offset $\Delta \mathbf{S}$, we can optimize the MSE loss as:
\begin{equation}\label{equ-loss-mse}
\mathcal{L}^{mse} = \|  \mathbf{S}^{*} - \mathbf{S} - \Delta\mathbf{S} \|^{2}.
\end{equation}

\begin{figure}[!htbp]
\begin{center}
  \subfigure[Input]{\label{fig-loss-compare-a}
    \includegraphics[width=0.13\textwidth]{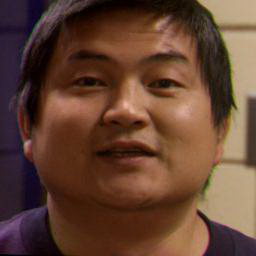}
  }
  \subfigure[Ground Truth]{\label{fig-loss-compare-b}
    \includegraphics[width=0.13\textwidth]{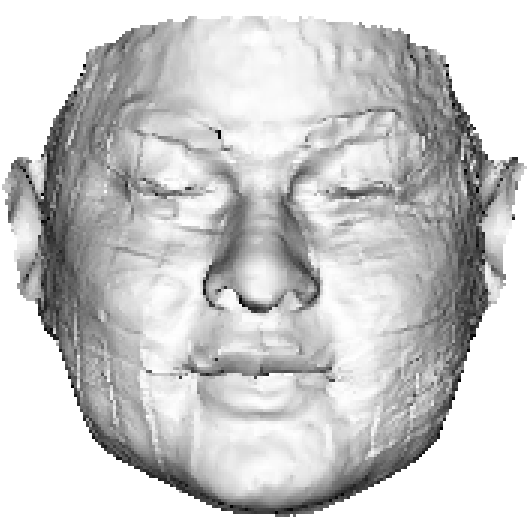}
  }
  \subfigure[3DMM]{\label{fig-loss-compare-c}
    \includegraphics[width=0.13\textwidth]{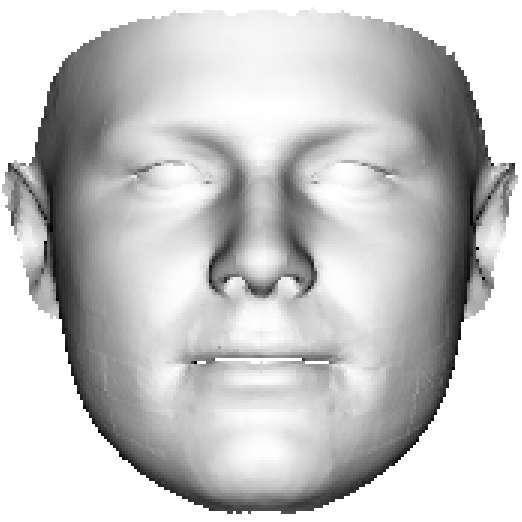}
  }\\
  \subfigure[MSE]{\label{fig-loss-compare-d}
    \includegraphics[width=0.13\textwidth]{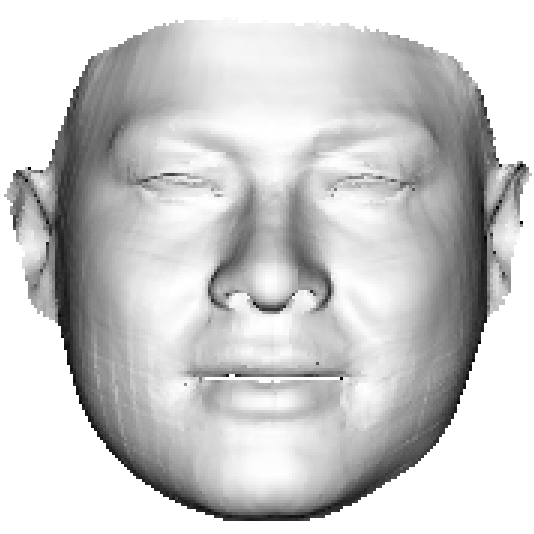}
  }
  \subfigure[PSD]{\label{fig-loss-compare-e}
    \includegraphics[width=0.13\textwidth]{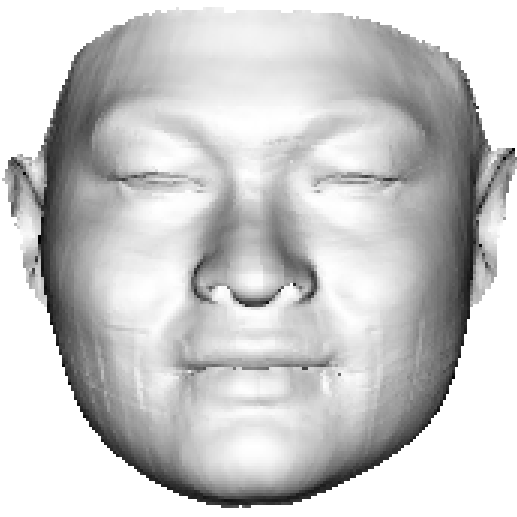}
  }
  \subfigure[VGD]{\label{fig-loss-compare-f}
    \includegraphics[width=0.13\textwidth]{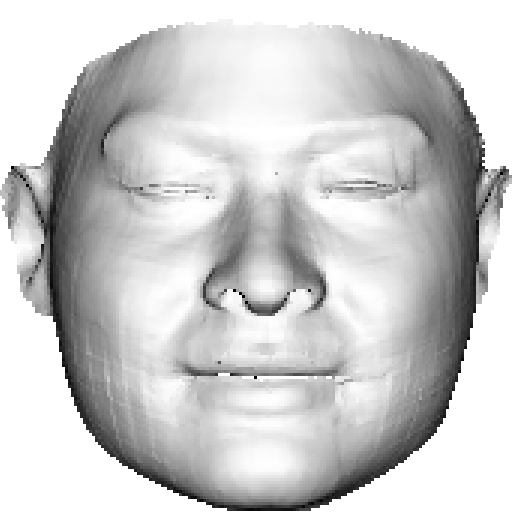}
  }
  \end{center}
  \caption{The reconstructed shapes by different losses. (a) The input image; (b) The ground-truth shape; (c) The 3DMM fitting result; (d) Mean Squared Error; (e) Plaster Sculpture Descriptor; (f) Visual-guided Loss. First, the visual difference between 3DMM and MSE is significant, especially for the nose and eyes. Second, PSD slightly refines the facial feature topology but the improvement is not obvious. Finally, VGD further reconstructs a sharper jaw that resembles the ground truth.}
  \label{fig-loss-compare}
\end{figure}

In Section~\ref{sec:introduction}, we argue that it is the limited representation power of 3DMM and the lack of the 3D ground truth that accounts for the model-like reconstruction, as shown in Fig.~\ref{fig-loss-compare-c}. However, with the shape reconstruction network trained on sufficient data, the reconstruction results are still not visually discriminative when considering the MSE loss only, as shown in Fig.~\ref{fig-loss-compare-d}. The main reason is that the vertex coordinate error does not account for how humans observe a 3D object. Therefore, a powerful 3D feature accounting for the visual effect, i.e., the $\mathcal{F}(\cdot)$ in Eqn.~\ref{eqn-task}, is crucial for loss functions. In this section, we propose a new \textbf{P}laster \textbf{S}culpture \textbf{D}escriptor (\textbf{PSD}) to model the visual effect and a \textbf{V}isual-\textbf{G}uided \textbf{D}istance (\textbf{VGD}) loss to supervise the network training.

\subsection{Plaster Sculpture Descriptor}
In 3D object retrieval, the light field descriptor~\cite{chen2003visual} is widely computed for silhouette images of 3D shapes. Besides, recent analysis-by-synthesis 3D face reconstruction methods~\cite{tran2019learning,zhou2019dense,gao2020semi} optimize  face images generated by a facial appearance model. Inspired by these two achievements, we consider that the images rendered by a 3D face can be utilized to model its visual effect, and explore a \textbf{P}laster \textbf{S}culpture \textbf{D}escriptor (\textbf{PSD}) to measure the reconstruction quality directly by how we see it. As shown in Fig.~\ref{fig-overview-c}, a 3D shape is considered as a plaster sculpture with all-white vertex color and \textbf{orthogonal light} on it. During training, the 3D shape is rendered at $5$ views, whose pitch and yaw angles are $(0^{\circ},0^{\circ})$, $(0^{\circ},90^{\circ})$, $(0^{\circ},-90^{\circ})$, $(30^{\circ},0^{\circ})$ and $(-30^{\circ},0^{\circ})$, and the L2 distances of the rendered images between the output and the ground-truth 3D shapes are formulated as the visual-effect distance:
\begin{equation}\label{eqn-loss-psd}
\mathcal{D}^{psd} = \sum_{v=1}^{5}  \|  \mathcal{R}(\mathbf{R}_{v}*(\mathbf{S} + \Delta \mathbf{S}), \mathbf{T}^{w}) -  \mathcal{R}(\mathbf{R}_{v}*\mathbf{S}^{*}, \mathbf{T}^{w}) \|_{2},
\end{equation}
where $v$ is the view index, $\mathcal{R}(\cdot, \cdot)$ is the renderer whose input is 3D shape and texture, $\mathbf{R}_{v}$ is the rotation matrix corresponding to each view, and $\mathbf{T}^{w}$ is the all-white texture under orthogonal light. If we employ a differentiable renderer~\cite{ravi2020accelerating}, the $\mathcal{D}^{psd}$ can be directly regarded as a loss function. As shown in Fig.~\ref{fig-loss-compare-e}, the facial features recovered by PSD are more similar to the ground truth, which is crucial in visual evaluation.

\begin{figure}[htb]	
\begin{center}
\includegraphics[width=0.48\textwidth]{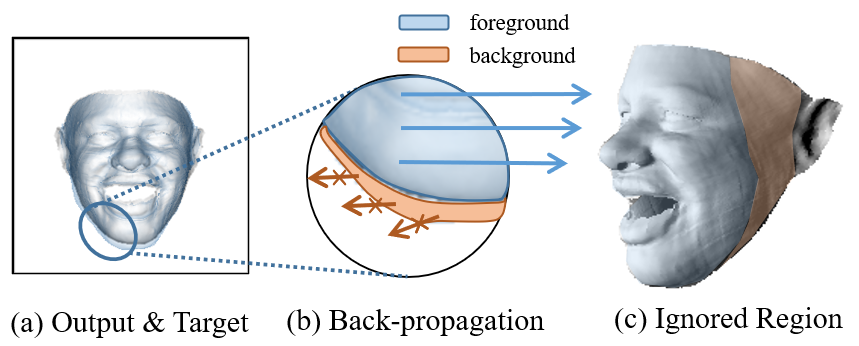}	\
\end{center}					
\caption{The defect of Plaster Sculpture Descriptor (PSD) when regarded as a loss function. (a) The output 3D shape overlapped with its target. (b) The back-propagation of PSD. If an output pixel is located on the background, its corresponding vertex does not have back-propagated signals. (c) The trained (blue) and ignored (red) regions. The face contour is usually located on the background and loses the supervision.}
\label{fig-PSD-defect}
\end{figure}

\subsection{Visual-guided Distance Loss}\label{sec-vgd}
\begin{figure}[htb]
\begin{center}
\includegraphics[width=0.42\textwidth]{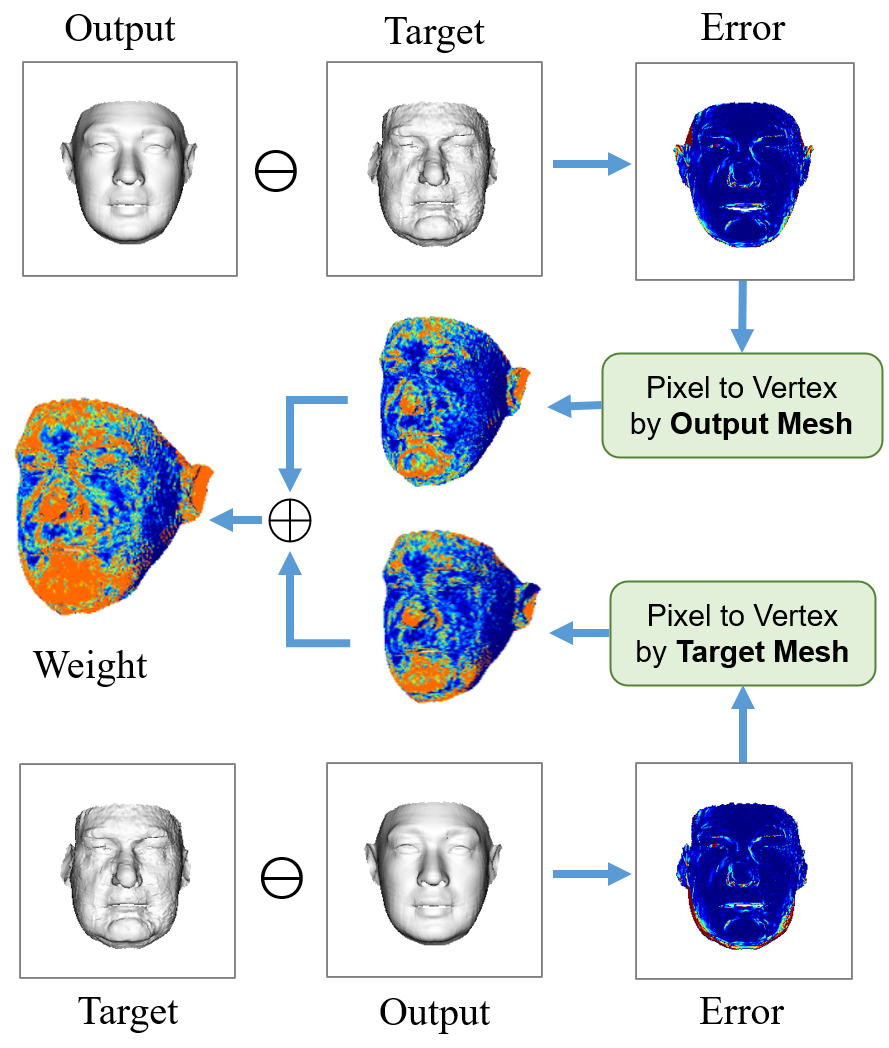}
\end{center}
\caption{Illustration of visual-guided weights calculation in one view. The output-to-target and the target-to-output PSD errors are traced back to the output and the target 3D mesh, respectively. The added errors are considered the vertex weights measuring the contribution to the visual effect.}
\label{fig-loss-VGD}
\end{figure}

Although plaster sculpture descriptor measures how the shapes are observed by humans, there is an intrinsic problem on the face contour due to the defect of the differentiable renderer. As shown in Fig.~\ref{fig-PSD-defect}, on the PSD error map, the differentiable renderer back-propagates the gradients of pixels to the corresponding vertices. Therefore, only the outputted vertices that located within the target face region have the back-propagated signals, and those on the background are not trained, leading to unsatisfactory contour reconstruction. To tackle this issue, we propose a \textbf{V}isual-\textbf{G}uided \textbf{D}istance (\textbf{VGD}) loss which considers the visual-effect distance as the vertex weights rather than the optimization target. The overview of VGD is shown in Fig.~\ref{fig-loss-VGD}, whose insight is that, given the ground-truth position of each vertex, we only need to find which vertices dominate the visual effect. Specifically, we calculate the output-to-target and the target-to-output PSD errors simultaneously, on either of which the poorly fitted face contour inevitably brings large values. Then, we employ pixel-to-vertex mapping to retrace the pixel errors to vertices, and add the vertex errors from both the output and the target. Finally, the accumulated vertex errors from all the views are regarded as the vertex weights for the MSE loss:
\begin{align}\label{eqn-loss-VGD}
\mathcal{L}^{vgd} &= \mathbf{W}^{psd} \odot \|  \mathbf{S}^{*} - \mathbf{S} - \Delta\mathbf{S} \|^{2},  \\
\mathbf{W}^{psd}  &=\sum_{v=1}^{5}[ \mathcal{R}_{ \mathbf{\mathbf{S} + \Delta\mathbf{S}}}^{-1}(\mathcal{D}^{psd}_v) + \mathcal{R}_{ \mathbf{S^*}}^{-1}(\mathcal{D}^{psd}_v)], \notag
\end{align}
where $\mathbf{W}^{psd}$ represents the vertex weights, $\mathcal{R}_{ \mathbf{S}}^{-1}(\cdot)$ denotes the inverse rendering function that maps pixel errors to the vertices according to the 3D face $\mathbf{S}$, and $\mathcal{D}^{psd}_v$ denotes the PSD error map in the $v$th view. As shown in Fig.~\ref{fig-loss-compare-f}, the visual-guided weights enable the network to focus on the visual discriminative regions, such as the face contour and the rough regions.

\section{Data Construction}\label{sec-data}
It is very tedious to acquire complete and high-precision 3D faces in real circumstances. The raw scans must be captured under well-controlled conditions and registered to a face template through laborious hand-labeling. To collect large-scale 3D data, we construct the 3D ground truth from single-view RGB-D images, which is more promising for massive collection, especially considering the rapid development of hand-held depth cameras such iPhone X.

\subsection{RGB-D Registration} \label{sec-register}
\begin{figure}[htb]
  \subfigure{
  \label{fig-icp-a}}
  \subfigure{
  \label{fig-icp-b}}
  \subfigure{
  \label{fig-icp-c}}
	\begin{center}
		\includegraphics[width=0.48\textwidth]{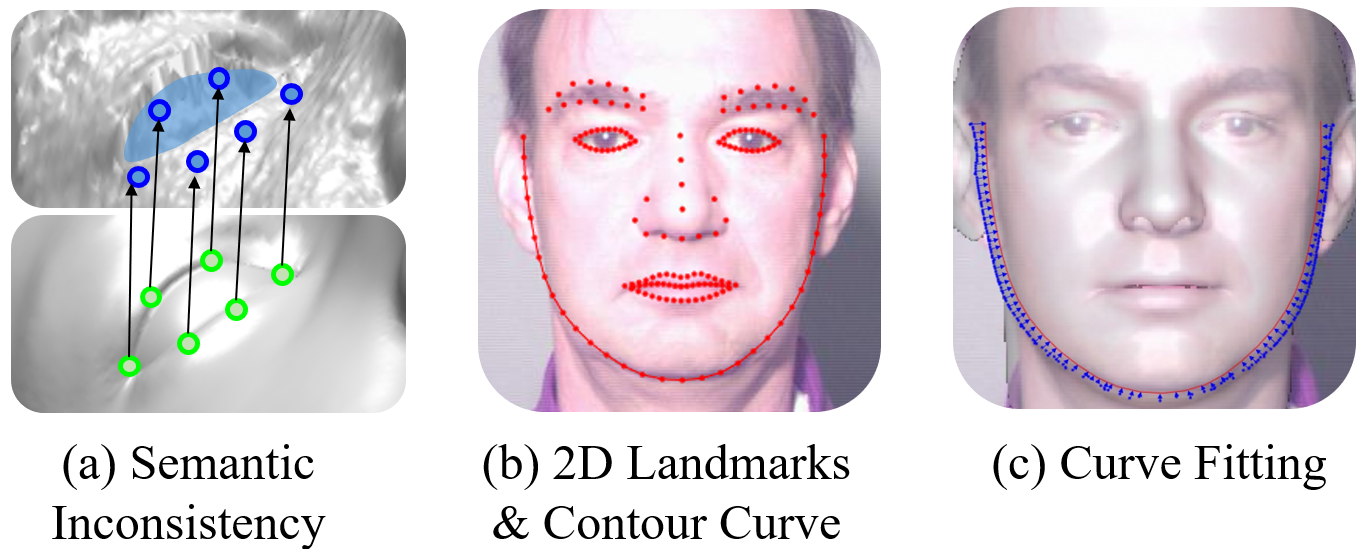}
	\end{center}
	\caption{RGB-D registration. (a) ICP only cannot guarantee semantic consistency. The example shows that the closest points on the target found by the eye-contour vertices are located outside the eye. (b) The landmarks on the facial-feature edge and the face contour curve. (c) The 3D contour vertices are fitted to the 2D contour curve.}
	\label{fig-icp}
\end{figure}

The training data requires face images and their corresponding topology-uniform 3D shapes. Thus, the first task is registering a template to the depth images, where Iterative Closest Point (ICP) method is commonly adopted. However, plausible registered results are not sound enough to be the supervision of neural network training. Semantic consistency, i.e., each vertex has the same semantic meaning across 3D faces is a critical criterion. As shown in Fig.~\ref{fig-icp-a}, semantically consistent registration cannot be achieved by finding the closest point in $(x,y,z)$ only. Considering that our registration target is RGB-D data, whose texture contains discriminative features for vertex matching, we introduce two terms on the RGB channels to provide more constraints in the popular optimal nonrigid-ICP loss function~\cite{Alpha10_amberg2007optimal}.

The first term comes from the edge. As shown in Fig.~\ref{fig-icp-b}, a series of landmarks, which are dense enough to finely enclose the eyes, eyebrows and mouth, are detected to represent the edges on the facial features. The constraint is formulated as:
\begin{equation}\label{equ-icp-landmark}
E_{edge} = \left\|\left[
\begin{array}{cccc}
a^{11} & a^{12} &  a^{13} & a^{14}\\
a^{21} & a^{22} &  a^{23} & a^{24}\\
\end{array}
\right]
\left[
\begin{array}{c}
x^{3D}\\
y^{3D}\\
z^{3D}\\
1\\
\end{array}
\right]
-
\left[
\begin{array}{c}
x^{2D}\\
y^{2D}\\
\end{array}
\right]\right\|,
\end{equation}
where $
\left[
\begin{array}{cccc}
a^{11} & a^{12} &  a^{13} & a^{14}\\
a^{21} & a^{22} &  a^{23} & a^{24}\\
\end{array}
\right]
$
is the first and second rows of the $3 \times 4$ affine transformation matrix of each vertex, which should be optimized, $(x^{3D}, y^{3D}, z^{3D})$ is an edge 3D landmark and $(x^{2D}, y^{2D})$ is the corresponding 2D landmark detected on the image.

The second term aims to regularize the face contour. However, it is unreliable to optimize the vertex-to-landmark distances as in Eqn.~\ref{equ-icp-landmark}, since there is no strict correspondence between 3D and 2D contour landmarks~\cite{zhu2015high}. To utilize contour landmarks, we perform curve fitting rather than landmark matching to reduce the semantic ambiguity. As shown in Fig.~\ref{fig-icp-c}, the 2D contour landmarks are sequentially connected to a contour curve. Then, the 3D contour vertices found by landmark marching~\cite{zhu2015high} are fitted to the 2D curve by:
\begin{align}\label{equ-icp-curve}
E&_{cont}=\left\|\left[
\begin{array}{cccc}
a^{11} & a^{12} &  a^{13} & a^{14}\\
a^{21} & a^{22} &  a^{23} & a^{24}\\
\end{array}
\right]
\left[
\begin{array}{c}
x^{3D}\\
y^{3D}\\
z^{3D}\\
1\\
\end{array}
\right]
-
\left[
\begin{array}{c}
x^{2D}_{c}\\
y^{2D}_{c}\\
\end{array}
\right]\right\|, \\
(x&^{2D}_{c}, y^{2D}_{c}) = \arg \min\limits_{(x, y)} \|(x, y) - (x^{3D}, y^{3D})\|, \forall (x,y) \in \mathbb{C}, \notag
\end{align}
where $(x^{2D}_{c}, y^{2D}_{c})$ is the 2D-closest point on the curve $\mathbb{C}$ to the 3D vertex $(x^{3D}, y^{3D}, z^{3D})$. During ICP registration, Eqn.~\ref{equ-icp-landmark} and Eqn.~\ref{equ-icp-curve} are employed as additional terms constraining the first two rows of affine transformations, and the third row is optimized separately by traditional ICP constraints~\cite{Alpha10_amberg2007optimal}.

Given the registered face $\mathbf{V}_{regist}$, we further disentangle rigid and non-rigid transformations by:
\begin{equation}\label{equ-disentangle_rigid}
\arg \min\limits_{\mathbf{S}, \mathbf{R}, f, \mathbf{t}_{3d}} \| \mathbf{V}_{regist} - (f * \mathbf{R}*\mathbf{S} +\mathbf{t}_{3d})\|,
\end{equation}
where $(f, \mathbf{R}, \mathbf{t}_{3d})$ are the rigid transformation parameters and the optimized shape is regarded as the ground-truth shape $\mathbf{S}^{*}$. The difference between the ground-truth shape and the 3DMM-fitted shape $\Delta\mathbf{S}=\mathbf{S}^{*} - \mathbf{\overline{S}} - \mathbf{A}_{id}\bm{\alpha}_{id} - \mathbf{A}_{exp}\bm{\alpha}_{exp}$ will be the target of the neural network training.

\subsection{Full-view Augmentation}\label{sec-augmentation}
\begin{figure}[htb]
\subfigure{
  \label{fig-profiling-a}}
  \subfigure{
  \label{fig-profiling-b}}
  \subfigure{
  \label{fig-profiling-c}}
  \subfigure{
  \label{fig-profiling-d}}
  \subfigure{
  \label{fig-profiling-e}}
  \subfigure{
  \label{fig-profiling-f}}
  \subfigure{
  \label{fig-profiling-g}}

  \begin{center}
     \includegraphics[width=0.48\textwidth]{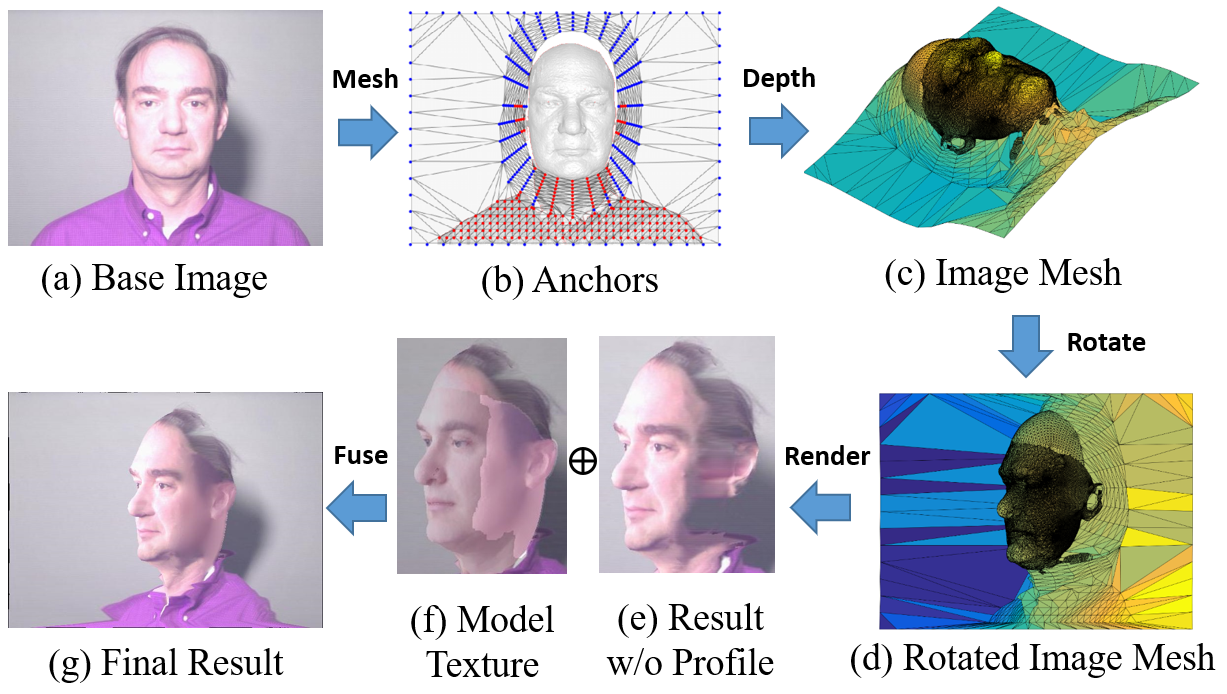}
  \end{center}
  \caption{An overview of full-view augmentation. (a) The base image; (b) The original depth channel and the anchors on the background. The red anchors have depth values and the blue anchors are located on the hollow, they have different constraints in Eqn.~\ref{equ-anchor-adjust}; (c) The completed depth channel; (d) The rotated 3D mesh; (e) Rendering with image pixels and model texture; (f) The augmentation result.}
  \label{fig-profiling}
\end{figure}

Currently, most of the public RGB-D images are frontal faces~\cite{CASIA,zhang2014bp4d,phillips2005overview}, leading to the risk of poor pose robustness. Although RGB-D data can be rendered to other views, the results suffer from large artifacts due to the incomplete depth channel and face texture. To address this issue, a full-view augmentation method is proposed specifically for RGB-D data. Based on the face profiling method~\cite{zhu2019face}, we complete the depth channel for the whole image space, where the depth on the face region comes directly from the registered 3D face and the depth on the background is coarsely estimated by some anchors, as shown in Fig.~\ref{fig-profiling-b}. Specifically, these anchors $(x_{i},y_{i})$ are triangulated to a background mesh and their depth values $d_{i}$ are regularized by a depth-channel constraint and a smoothness constraint:
\begin{equation}\label{equ-anchor-adjust}
\begin{aligned}
   &\sum_{i} Mask(x_{i},y_{i}) \| d_{i} - Depth(x_{i},y_{i}) \| + \\
   &\sum_{i}\sum_{j} Connect(i,j) \| d_{i} - d_{j} \|,
\end{aligned}
\end{equation}
where $Depth(x,y)$ is the depth channel of the RGB-D data, $Mask(x,y)$ indicates whether $(x,y)$ is hollow, and $Connect(i,j)$ indicates whether two anchors are connected by the background mesh. With the complete depth channel (Fig.~\ref{fig-profiling-c}), the RGB-D can be rotated (Fig.~\ref{fig-profiling-d}) and rendered (Fig.~\ref{fig-profiling-e}) to any views. More implemental details are provided in the supplemental materials.

Different from face profiling~\cite{zhu2019face}, whose purpose is enlarging medium poses to large poses, our method aims to augment frontal faces. However, the main drawback of the rotate-and-render strategy is that, when rotating from frontal faces, there are serious artifacts on the side face due to the incomplete face texture, as shown in Fig.~\ref{fig-profiling-e}. In this work, we employ the texture and illumination model in 3DMM as a strong prior to refine the artifacts. Specifically, the face texture is modeled by PCA~\cite{Paysan-AVSS-09}:
\begin{equation}\label{equ-3dmm-tex}
\mathbf{T}=\mathbf{\overline{T}} + \mathbf{B}\bm{\beta},
\end{equation}
where $\mathbf{\overline{T}}$ is the mean texture, $\mathbf{B}$ is the principle axis of the texture and $\bm{\beta}$ is the texture parameter. Given 3D shape $\mathbf{V}$ and texture $\mathbf{T}$, the Phong illumination model is used to produce the color of each vertex~\cite{1_blanz2003face}:
\begin{equation}\label{equ-3dmm-illum}
C_{i}(\mathbf{p}_{tex})= \mathbf{Amb}*\mathbf{T}_{i} + \mathbf{Dir}*\mathbf{T}_{i}*\langle \mathbf{n}_{i},\mathbf{l} \rangle + k_{s}\cdot \mathbf{Dir}\langle \mathbf{r}_{i}, \mathbf{ve} \rangle^{\nu},
\end{equation}
where $C_{i}$ is the RGB of the $i$th vertex, the diagonal matrix $\mathbf{Amb}$ is the ambient light, the diagonal matrix $\mathbf{Dir}$ is the parallel light from direction $\mathbf{l}$, $\mathbf{n}_{i}$ is the normal direction of the $i$th vertex, $k_{s}$ is the specular reflectance, $\mathbf{ve}$ is the viewing direction, $\nu$ controls the angular distribution of the specular
reflection and $\mathbf{r}_{i}=2\cdot\langle \mathbf{n}_{i},\mathbf{l} \rangle\mathbf{n}_{i}-\mathbf{l}$ is the direction of maximum specular reflection. The collection of texture parameters is $\mathbf{p}_{tex} = [\bm{\beta}, \mathbf{Amb}, \mathbf{Dir}, \mathbf{l}, k_{s}, \nu]$. Given the ground-truth 3D shape $\mathbf{V}_{regist}$, the texture parameters $\mathbf{p}_{tex}$ can be estimated by the analysis-by-synthesis manner:
\begin{equation}\label{equ-loss-pose-aug}
\arg ~\min\limits_{\mathbf{p}_{tex}}  ~ \| \mathbf{I}(\mathbf{V}_{regist}) - C(\mathbf{p}_{tex}) \|,
\end{equation}
where $\mathbf{I}(\mathbf{V})$ is the image pixels at the vertex positions. The optimized result $C(\mathbf{p}_{tex})$ is the face texture, as shown in Fig.~\ref{fig-profiling-f}. With the estimated model texture, we render the 3D face with both the image pixels and the model texture, as shown in Fig.~\ref{fig-profiling-e} and Fig.~\ref{fig-profiling-f}. Then, we inpaint the self-occluded side face with the model texture through Poisson editing, as shown in Fig.~\ref{fig-profiling-g}. It can be seen that the model texture is realistic enough to inpaint the smoothly textured side face.

\subsection{Shape Transformation}\label{sec-shape-transfer}
\begin{figure}[htb]
\subfigure{
  \label{fig-shape-transfer-a}}
  \subfigure{
  \label{fig-shape-transfer-b}}
  \subfigure{
  \label{fig-shape-transfer-c}}
  \subfigure{
  \label{fig-shape-transfer-d}}
  \subfigure{
  \label{fig-shape-transfer-e}}
  \subfigure{
  \label{fig-shape-transfer-f}}

  \begin{center}
     \includegraphics[width=0.5\textwidth]{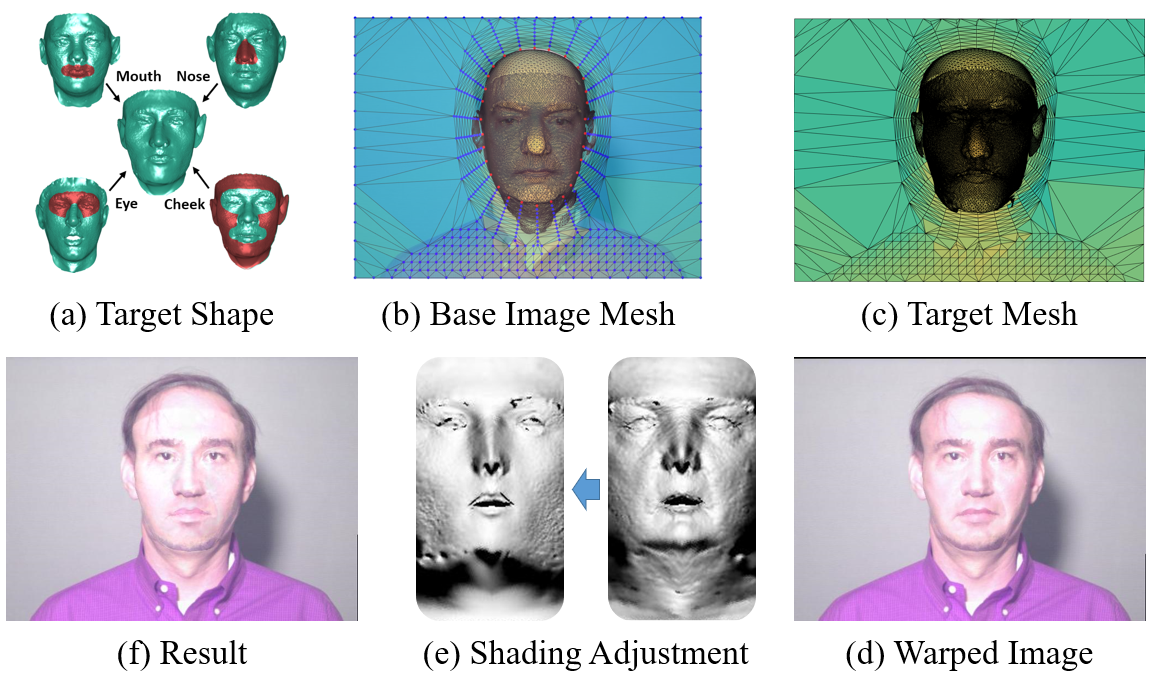}
  \end{center}
  \caption{An overview of shape transformation. (a) Target shape generation by fusing facial features of different identities; (b) The original image and its 3D mesh;  (c) The target 3D mesh; (d) The result of shape transformation and background warping; (e) The shading adjustment; (f) The final result.}
  \label{fig-shape-transfer}
\end{figure}

In addition to pose variations, shape variations covered by the training data are also important since the personalized shape is the main goal. Unfortunately, existing data does not contain sufficient identities due to demanding data collection. In this section, we propose a shape augmentation method to transform the underlying 3D shape of a face image and refine the face appearance accordingly. In specific, we first construct the target shape to transform, whose eye, nose, mouth, and cheek come from different faces in the datasets, as shown in Fig.~\ref{fig-shape-transfer-a}. Second, the base image is tuned into a 3D mesh following the same process in the full-view augmentation, as shown in Fig.~\ref{fig-shape-transfer-b}. Third, we replace the 3D shape and warp the background to adjust the new face. Specifically, the background anchors are adjusted by minimizing the following function:
\begin{equation}\label{equ-anchor-shape_transfer}
\begin{aligned}
  &\sum_{i} FaceContour(i) * (\| x^{t}_{i} - x^{s}_{i} \| +  \| y^{t}_{i} - y^{s}_{i} \|)+ \\
  &\sum_{i}\sum_{j} Connect(i,j) * (\| (x^{t}_{i} - x^{t}_{j}) - (x^{s}_{i} - x^{s}_{j}) \| \\
  &+\| (y^{t}_{i} - y^{t}_{j}) - (y^{s}_{i} - y^{s}_{j}) \|),
  \end{aligned}
\end{equation}
where $(x^{s}_{i},y^{s}_{i})$ is the anchor position on the source image, $(x^{t}_{i},y^{t}_{i})$ is its target position on the augmented image, $FaceContour(i)$ indicates whether anchor $i$ is located on the face contour (the red points in Fig.~\ref{fig-shape-transfer-b}), and $Connect(i,j)$ indicates whether two anchors are connected by the background mesh. Afterwards, we render the warped 3D mesh and obtain the shape-transformed result, whose ground-truth 3D shape is the target 3D shape, see Fig.~\ref{fig-shape-transfer-d}.

In addition to image pixel warping, shading adjustment is also important according to the shape-from-shading theory~\cite{kemelmacher20113d}, which can be achieved by the illumination model the same as Eqn.~\ref{equ-3dmm-illum}:
\begin{equation}\label{equ-3dmm-illum-shape-transfer}
\mathbf{C}_{i}^{t}= \mathbf{Amb}*\mathbf{T}_{i} + \mathbf{Dir}*\mathbf{T}_{i}*\langle \mathbf{n}^{t}_{i},\mathbf{l} \rangle + k_{s}\cdot \mathbf{Dir}\langle \mathbf{r}^{t}_{i}, \mathbf{ve} \rangle^{\nu},
\end{equation}
where all the parameters except $\mathbf{n}^{t}$ and $\mathbf{r}^{t}$ are from the source image, and $\mathbf{n}^{t}$ and $\mathbf{r}^{t}$, which account for shading, are from the target shape. Finally, we change the facial pixels to $\mathbf{C}^{t}$ and obtain the final result of shape transformation, see Fig.~\ref{fig-shape-transfer-f}. We provide more implemental details in the supplemental materials.

\section{Benchmarks}
\begin{figure*}[htb]
\subfigure{
  \label{fig-dataset-a}}
  \subfigure{
  \label{fig-dataset-b}}
  \subfigure{
  \label{fig-dataset-c}}
  \subfigure{
  \label{fig-dataset-d}}
  \begin{center}
     \includegraphics[width=0.96\textwidth]{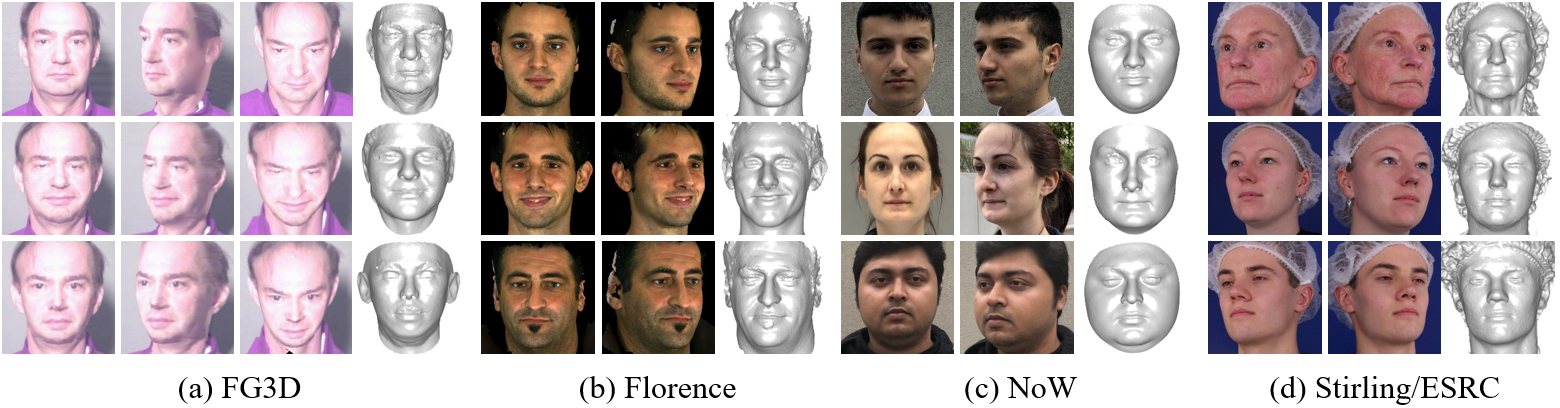}
  \end{center}
  \caption{Selected examples from the datasets employed in the experiments. Each face image has a 3D ground-truth shape.}
  \label{fig-shape-transfer}
\end{figure*}

\subsection{Dataset}
{\bfseries Fine-Grained 3D Face (FG3D)} is constructed from three datasets, FRGC, BP4D, and CASIA-3D. FRGC~\cite{phillips2005overview} includes $4,950$ samples, each of which has a face image and a 3D scan in full correspondence. BP4D~\cite{zhang2014bp4d} contains $328$ 2D+3D videos from $41$ subjects, where $3,376$ frames are randomly selected in total. CASIA-3D~\cite{CASIA} consists of $4,624$ scans of $123$ persons and the non-frontal faces are filtered out.

We divide $90\%$ of subjects as the training set \textbf{FG3D-train} and the remaining $10\%$ of subjects as the testing set \textbf{FG3D-test}. For the training set, we perform shape transformation as in Section~\ref{sec-shape-transfer} $4$ times to augment shape variations. Then, the out-of-plane pose variations are augmented by full-view augmentation as in Section~\ref{sec-augmentation}, where the $yaw$ angle is enlarged at steps of $15^{\circ}$ until $50^{\circ}$ and the $pitch$ angle is enlarged by $15^{\circ}$ and $-25^{\circ}$, generating $474k$ samples in total. For the testing set, we only perform full-view augmentation and generate $12k$ samples. Besides, we manually delete the bad registration results in the FG3D-test manually for better evaluation. This dataset is employed for the model training and the performance analysis of our proposal. Some examples are shown in Fig.~\ref{fig-dataset-a}.

{\bfseries Florence}~\cite{bagdanov2011florence} is a 3D face dataset containing 53 subjects with 3D meshes acquired from a structured-light scanning system. Based on the protocol in~\cite{16_feng2018joint}, each subject is rendered at pitches of $-20^{\circ}$, $0^{\circ}$, $20^{\circ}$ and yaws of $-45^{\circ}$ to $45^{\circ}$ at the step of $15^{\circ}$, as shown in Fig.~\ref{fig-dataset-b}. Besides, we register each 3D mesh with hand-labeled landmarks and carefully check the registration results. This dataset is used for cross-dataset evaluation to demonstrate the generalization.

{\bfseries NoW}~\cite{Sanyal_2019_CVPR} contains 2054 face images of $100$ subjects, which are split into an open validation set ($20$ subjects) and a private test set ($80$ subjects). Each subject has face images with different poses and expressions, and one neutral 3D face scan for reference. Under the original protocol, the reconstruction results should be disentangled from the expression to compare with the neutral 3D scan, which is not consistent with our goal. Therefore, we only select the neural-expression images of the validation set, as shown in Fig.~\ref{fig-dataset-c}, for cross evaluation.

{\bfseries Stirling/ESRC}~\cite{Stirling/ESRC} is a 3D face dataset with $134$ subjects ($64$ males and $70$ females). Each subject has a 3D face scan in a neutral expression and two corresponding face images in $yaw=\pm 45^{\circ}$, which are captured by the Di3D camera system simultaneously. In the experiments, all subjects are employed in the cross-dataset evaluation. Some examples are shown in Fig.~\ref{fig-dataset-d}.

\subsection{Evaluation Metric}
It is still an open problem to determine the proper measurement of 3D shape accuracy. Errors calculated after projection~\cite{16_feng2018joint, zhu2019face, deng2019accurate} mostly account for pose accuracy since pose is the dominant factor of vertex positions~\cite{zhu2019face}. To highlight shape accuracy, which is our main purpose, we normalize pose before error calculation. Specifically, we register a face template to the ground-truth 3D face as in Section~\ref{sec-register}, finding the vertex correspondence $(k, k^{t})$, where $k$ is the vertex index on the face template and $k^{t}$ is that on the ground truth. We also note $k \in \mathcal{C}$ if $k$ belongs to the face region and the correspondence $(k, k^{t})$ is reliable (the spatial and normal distances are below thresholds). For each reconstruction result $\mathbf{V}$ and the ground truth $\mathbf{V}^{*}$, we estimate a rigid transformation:
\begin{align}
\arg \min\limits_{f, \mathbf{R}, \mathbf{t}_{3d}} \sum_{k \in \mathcal{C}}\| (f * \mathbf{R} * \mathbf{v}_{k} + \mathbf{t}_{3d}) -  \mathbf{v}^{*}_{k^{t}}\|,
\end{align}
where $(f, \mathbf{R}, \mathbf{t}_{3d})$ are the rigid transformation parameters for pose alignment. Based on this rigid alignment, we utilize the widely applied Normalized Mean Error (NME) to measure the reconstruction error:
\begin{align}\label{eqn-DAEM}
NME = \frac{1}{K}\sum_{k=1}^{K}\frac{||(f * \mathbf{R} * \mathbf{v}_{k} + \mathbf{t}_{3d}) - \mathbf{v}^{*}_{k^{t}}||}{d},
\end{align}
where $K$ is the number of vertices, $\mathbf{v}_{k}$ is a vertex on the reconstructed face, $\mathbf{v}^{*}_{k^{t}}$ is the target of $\mathbf{v}_{k}$, and $d$ is the 3D outer interocular distance. This error is adopted in the performance analysis on the FG3D-test, where we trust the registration results.

We also employ a Densely Aligned Chamfer Error (DACE) to measure the distances of the closest points:
\begin{align}
DACE = \frac{1}{\mathcal{N}(C)}\sum_{k \in \mathcal{C}}\frac{||(f * \mathbf{R} * \mathbf{v}_{k} + \mathbf{t}_{3d}) - \mathbf{v}^{*}_{k^{nn}}||}{d},
\end{align}
where $\mathbf{v}^{*}_{k^{nn}}$ is the nearest neighbor of $\mathbf{v}_{k}$ on the raw 3D scan after rigid alignment. We adopt this error in the comparison experiments on Florence, NoW and Stirling/ESRC, where the raw 3D scans are employed as the target and different methods may have different topologies. It is worth noting that, only the vertex in set $\mathcal{C}$, which indicates that its ICP matching is reliable, participates in the error calculation, which filters out the errors on the noisy, hollow and occluded regions.

\section{Experiments}

\subsection{Implementation Details}
The initial 3D face is acquired by 3DDFA~\cite{3ddfa_cleardusk}. The backbone follows the network defined in PRNet~\cite{prnet-github} and the shortcuts are added to generate an hourglass network. The models are trained by the SGD optimizer with a starting learning rate of $0.1$, which is decayed by $0.1$ at epochs $30$, $35$ and $40$, and the model is trained for $45$ epochs. The training images are cropped by the bounding boxes of the initial 3DMM fitting results~\cite{3ddfa_cleardusk} and resized to $256 \times 256$ without any perturbation. The UV displacement map is also $256 \times 256$. In all the following experiments, the FG3D-train is employed as the training set. The testing is conducted on the FG3D-test for performance analysis with Normalized Mean Error (NME), and several other datasets for comparison with the state-of-the-art methods, where Densely Aligned Chamfer Error (DACE) is preferred.

\subsection{Network Structure Analysis}
In this section, we thoroughly analyze how the network structure benefits the reconstruction accuracy.

\subsubsection{Ablation Study on Network Properties}
\begin{table*}
\tabcolsep 7pt
	\caption{The Normalized Mean Error (NME) of different networks, evaluated on the FG3D-test with different yaw intervals.  ``Lossless'', ``Normalize'' and ``Concentrate'' are the three properties desired for network structures, which are discussed in Section~\ref{sec-network}, The symbol $\checkmark$ means the property is fulfilled. }
	\centering
		\begin{tabular}{cccccccc}
			\toprule[1.5pt]
		    Network Structure & Lossless & Normalize & Concentrate & $[0,15]$ & $[15,30]$ & $[30,45]$ & Mean \\
            \midrule[1pt]
            Initial Shape (3DMM) &   &   &   & 6.13 & 6.13 & 6.09 & 6.11\\
            \midrule[0.5pt]
			Camera View (CVN)~\cite{zhu2020beyond} & $\checkmark$ &  &  & 3.68 & 3.56 & 3.46 & 3.57\\
			Model View (MVN)~\cite{zhu2020beyond} &  & $\checkmark$ & $\checkmark$ & 3.54 & 3.47 & 3.46 & 3.49\\
            \textbf{Virtual Multiview (VMN)} & $\checkmark$ & $\checkmark$ & $\checkmark$ & \textbf{3.43} & \textbf{3.26} & \textbf{3.24}  & \textbf{3.32}\\
			\bottomrule[1.5pt]
		\end{tabular}
	\label{tab-network}
\end{table*}
In Section~\ref{sec-network}, we introduce three properties desired for network design: normalization, lossless and concentration, which require the network to normalize non-shape components, preserve the image information, and align receptive fields. To evaluate the benefits of the properties, we test two alternative networks proposed in our preliminary work~\cite{zhu2020beyond} that fulfill different levels of them. First, the Camera View Network (CVN)~\cite{zhu2020beyond} concatenates the original image and the Projected Normalized Coordinate Code (PNCC)~\cite{zhu2019face} encoded by the fitted 3DMM to regress the vertex displacements. The network fulfills the lossless property by preserving the original image. However, the pose variations are not normalized and the receptive field does not cover the most related region since the input and the output have different coordinate systems (image plane vs. UV plane). Thus, the normalization and the concentration properties are missed. Second, the Model View Network (MVN)~\cite{zhu2020beyond} performs explicit normalization by warping the image to the UV plane, making the receptive field cover the most related region. However, the normalization loses the 2D geometry and the external face regions of the original input. Thus, the lossless property is violated. Compared with them, our Virtual Multiview Network (VMN) fulfills all the principles by constructing a virtual multiview camera system and aligning intermediate deep features to the UV plane.

\begin{figure}[htb]
\begin{center}
  \subfigure[CVN]{
  \includegraphics[width=0.145\textwidth]{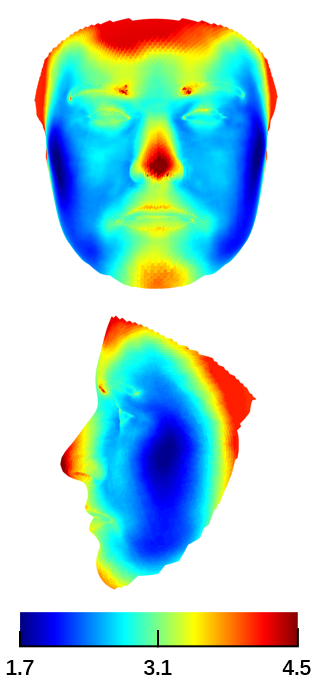}
  \label{fig-exp-shapediff-a}}
  \subfigure[MVN]{
  \includegraphics[width=0.145\textwidth]{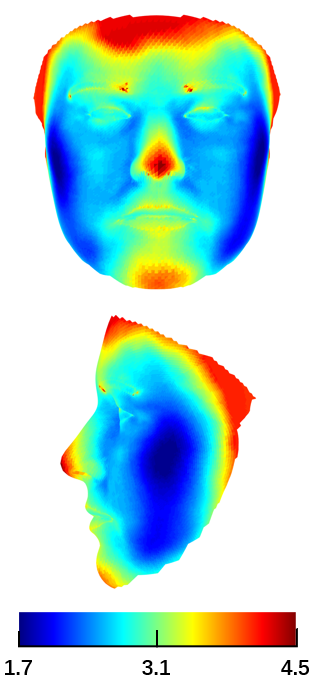}
  \label{fig-exp-shapediff-b}}
  \subfigure[VMN]{
  \includegraphics[width=0.145\textwidth]{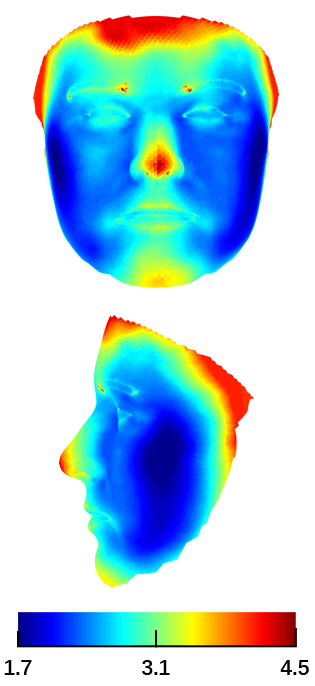}
  \label{fig-exp-shapediff-c}}
\end{center}
	\caption{The vertex-level NME of our method and other baselines, calculated by the mean of FG3D-test. Lower value indicates better accuracy. (a) Camera View Network (CVN), (b) Model View Network (MVN) and (c) our method (VMN).}
	\label{fig-exp-shapediff}
\end{figure}

We evaluate each network and the initial 3DMM fitting results in Table~\ref{tab-network}, and find that all the networks improve the initial shape. Besides, we observe that the more properties the network fulfills, the better its performance. The improvement achieved by replacing CVN with MVN reflects the effectiveness of explicitly normalizing non-shape components. Replacing MVN with VMN further promotes the performance, indicating the significance of preserving the original 2D geometry and the external face regions. Another interesting observation is that the medium poses, with yaw angles between $[30^{\circ},45^{\circ}]$, are the best for shape reconstruction, owing to the depth information for both side view and frontal view.

We further demonstrate the error at a more fine-grained vertex level, as shown in Fig.~\ref{fig-exp-shapediff}. First, there is a clear improvement from CVN to MVN, and further to VMN. Second, the improvements in the nose and jaw are the most obvious, followed by eye, mouth and apple cheeks. It is implied that personalized shape is mainly encoded in these regions.

\subsubsection{Input Views of Virtual Multiview Network}
\begin{figure}[htb]
\begin{center}
  \subfigure[Source]{\includegraphics[width=0.10\textwidth]{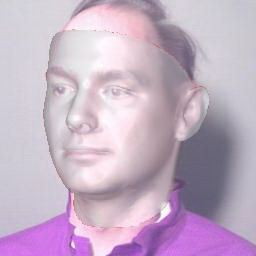}}
  \subfigure[$(0^{\circ},0^{\circ})$]{\includegraphics[width=0.10\textwidth]{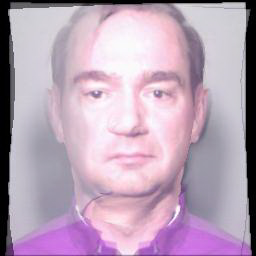}}
  \subfigure[$(15^{\circ},0^{\circ})$]{\includegraphics[width=0.10\textwidth]{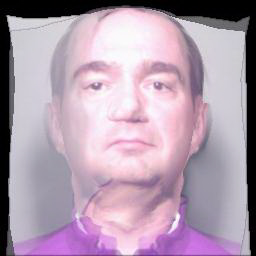}}
  \subfigure[$(-25^{\circ},0^{\circ})$]{\includegraphics[width=0.10\textwidth]{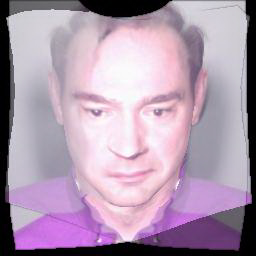}}\\
  \subfigure[$(0^{\circ},25^{\circ})$]{\includegraphics[width=0.10\textwidth]{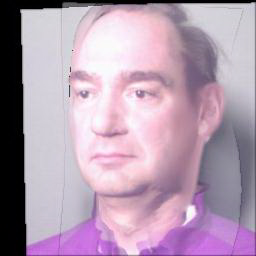}}
  \subfigure[$(0^{\circ},50^{\circ})$]{\includegraphics[width=0.10\textwidth]{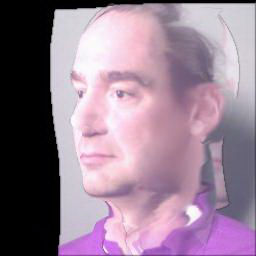}}
  \subfigure[$(0^{\circ},-25^{\circ})$]{\includegraphics[width=0.10\textwidth]{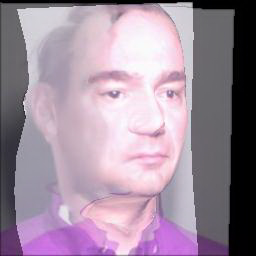}}
  \subfigure[$(0^{\circ},-50^{\circ})$]{\includegraphics[width=0.10\textwidth]{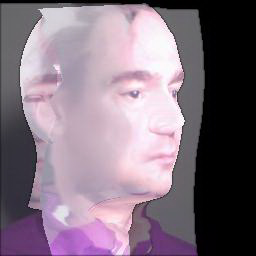}}
  \end{center}
  \caption{The $7$ virtual views of the input face. The angles in the bracket are $($pitch, yaw$)$.}
  \label{fig-cross-view}
\end{figure}

In virtual multiview network, the input image is rendered to several constant views to simulate a multi-camera system. In this section, we further analyze how many views are appropriate to describe the face shape, ranging from the frontal view only to multiple views with rich pose variations. Totally $7$ views are evaluated, whose $($pitch, yaw$)$ are $(0^{\circ},0^{\circ})$, $(15^{\circ},0^{\circ})$, $(-25^{\circ},0^{\circ})$, $(0^{\circ},25^{\circ})$, $(0^{\circ},50^{\circ})$, $(0^{\circ},-25^{\circ})$ and $(0^{\circ},-50^{\circ})$, as shown in Fig.~\ref{fig-cross-view}.

Table~\ref{tab-crossview} evaluates $4$ combinations of the $7$ views, where the original image is regarded as the baseline. First, the $1$-view does not perform well, even worse than the original input. It is worth noting that the $1$-view setting is equivalent to the frontalization strategy~\cite{Asthana-ICCV-2011,hassner2015effective,zhu2015high,chu20143d,masi2019face}, suggesting that, although widely applied in face recognition~\cite{masi2019face}, frontalization is not appropriate in shape reconstruction due to the loss of profile. Besides, the $3$-view outperforms the $1$-view with the provided yaw variations, verifying the benefits of side view information. The $5$-view further achieves the best result with the additional pitch variations. Finally, the $7$-view indicates that the yaw angles inverse to the original pose make little difference. Thus, we exclude them.

\begin{table}
\tabcolsep 5pt
	\caption{The Normalized Mean Error (NME) with different numbers of views, evaluated on all the samples of FG3D-test. The symbol $\checkmark$ means the view in (pitch, yaw) is employed.}
	\begin{center}
		\begin{tabular}{c c c c c c c  c c}
			\toprule[1.5pt]
            Pitch & $0^{\circ}$ & $15^{\circ}$ & $-25^{\circ}$ & $0^{\circ}$ & $0^{\circ}$ &  $0^{\circ}$ & $0^{\circ}$ & \multirow{2}{*}{NME}\\
            \cmidrule(lr){1-8}
            Yaw & $0^{\circ}$ & $0^{\circ}$ & $0^{\circ}$ & $25^{\circ}$ & $50^{\circ}$ &  $-25^{\circ}$ & $-50^{\circ}$ & \\
            \midrule[1pt]
            1-View & $\checkmark$  &   &   &  &   &   &   &  3.63\\
            3-View & $\checkmark$  &   &   & $\checkmark$  & $\checkmark$  &  &   & 3.40\\
			\textbf{5-View} & $\checkmark$  &  $\checkmark$ &  $\checkmark$ & $\checkmark$  & $\checkmark$  &  &   & \textbf{3.32}\\
		    7-View & $\checkmark$  &  $\checkmark$ &  $\checkmark$ & $\checkmark$ & $\checkmark$  &  $\checkmark$ & $\checkmark$  & 3.35\\
            \midrule[0.5pt]
            Source &   \multicolumn{7}{c}{same as input} & 3.60\\
			\bottomrule[1.5pt]
		\end{tabular}
	\end{center}
	\label{tab-crossview}
\end{table}

\subsubsection{Multiview Fusion in Many-to-one Hourglass}
\begin{figure}[htb]
    \begin{center}
     \includegraphics[width=0.48\textwidth]{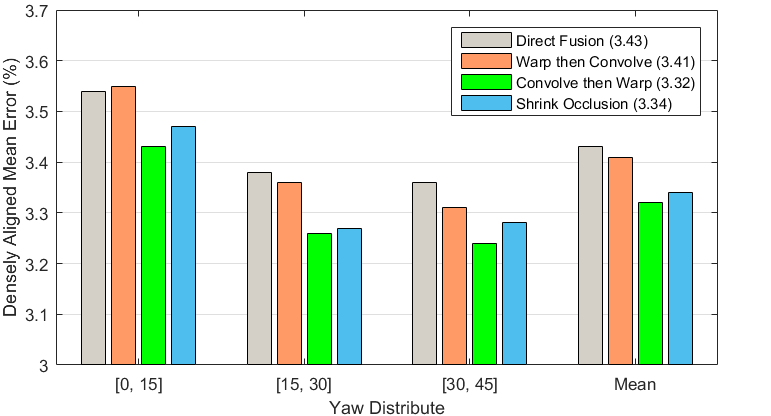}
	\end{center}
	\caption{The Normalized Mean Error (NME) of different multiview fusion methods, evaluated on the FG3D-test with different yaw intervals. The value in bracket is the mean NME of all the samples.}
	\label{fig-fusion}
\end{figure}

The virtual multiview network aims to fuse the features from $5$ views and output the personalized shape on the UV plane, where a many-to-one hourglass network is employed. Since the inputs and the output lie on different planes (image plane vs. UV plane), when the features are fused, e.g., concatenated in the middle and added in the symmetric layers, the fused features have different receptive fields and may degrade each other. We first employ direct feature fusion and achieve unsatisfactory performance, as shown in Figure~\ref{fig-fusion}. Second, we implement a weak alignment manner that first warps the input images to the UV plane and then performs convolution. Although slight improvements are achieved, this warp-then-convolve manner violates the lossless requirement since the face contour and the external face regions are lost. Furthermore, we evaluate the proposed method that first convolves the images and then warps the intermediate features to the UV plane before feature fusion. This convolve-then-warp manner achieves the best performance, showing the importance of preserving the information in the original image. Finally, we attempt to shrink the features on the occluded region as in~\cite{zhu2020beyond,zhu2019face}, but find subtle difference, which may be attributed to the self-occlusion inpainting manner in Sec.~\ref{sec-multiview-simulation}. To simplify the network, we do not employ the shrinking module.

\subsection{Data Augmentation Analysis}
\begin{figure}[htb]
\begin{center}
    \includegraphics[width=0.45\textwidth]{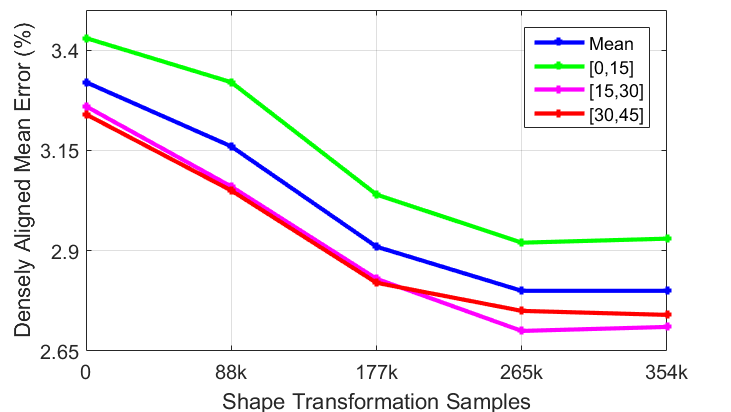}
  \end{center}
  \caption{The performance improvement as more shape-augmented samples are added in training, evaluated on the FG3D-test with different yaw intervals.}
  \label{fig-shape-transfer-adding}
\end{figure}

In Section~\ref{sec-data}, we augment face appearance in both pose and shape to provide adequate data for neural network training. In pose augmentation, we improve face profiling~\cite{zhu2019face} by completing the depth channel and inpainting the side face with a texture model. As shown in Table~\ref{tab-data-augment}, our full-view augmentation outperforms face profiling, especially in large poses, which is attributed to the artifacts exposed on the side face that face profiling cannot repair. Besides, by further incorporating shape transformation, the training data reaches $474$k samples and the error is greatly reduced by $15.7\%$. It can be seen that in high-fidelity reconstruction, where shape is mostly concerned, our shape transformation is a highly effective augmentation strategy. However, adding an overwhelming number of augmented samples may not benefit the performance as the shape-augmented samples are fake images. Therefore, we further add the shape-transformed samples progressively and observe how the performance changes. As shown in Fig.~\ref{fig-shape-transfer-adding}, the error reduction converges when approximately $265$k samples ($2.2$ times that of the original samples) are added.

\begin{table}[htb]
\tabcolsep 5pt
	\caption{The Normalized Mean Error (NME) of different data augmentation strategies, evaluated on the FG3D-test with different yaw intervals. ``Num'' is the number of training samples, ``Pose Aug'' indicates the full-view augmentation and ``Shape Aug'' refers to the shape transformation.}
	\begin{center}
		\begin{tabular}{cccccc}
			\toprule[1.5pt]
		    Augmentation & Num & $[0,15]$ & $[15,30]$ & $[30,45]$ & NME\\
            \midrule[1pt]
            Face Profiling~\cite{zhu2019face} &  120k  &  3.50   &    3.44    &     3.58   & 3.51\\
            Pose Aug &  120k   &   3.43 & 3.26 & 3.24  & 3.32 \\
            \textbf{Pose \& Shape Aug} &  \textbf{474k}   &  \textbf{2.93}  &   \textbf{2.71}   &    \textbf{2.74}   & \textbf{2.80}\\
			\bottomrule[1.5pt]
		\end{tabular}
	\end{center}
	\label{tab-data-augment}
\end{table}

\subsection{Loss Function Analysis}
In supervised learning, the loss function directly judges the reconstruction results according to the targets. Considering that the fine-grained geometry cannot be well captured by the traditional Mean Squared Error (MSE), we propose the Plaster Sculpture Descriptor (PSD) to model the visual effect and the Visual-Guided Distance (VGD) to supervise the training. We evaluate the loss functions by both Normalized Mean Error (NME) and Densely Aligned Chamfer Error (DACE), where the latter concerns more about the visual effect.

\begin{figure}[htb]
\begin{center}
    \includegraphics[width=0.48\textwidth]{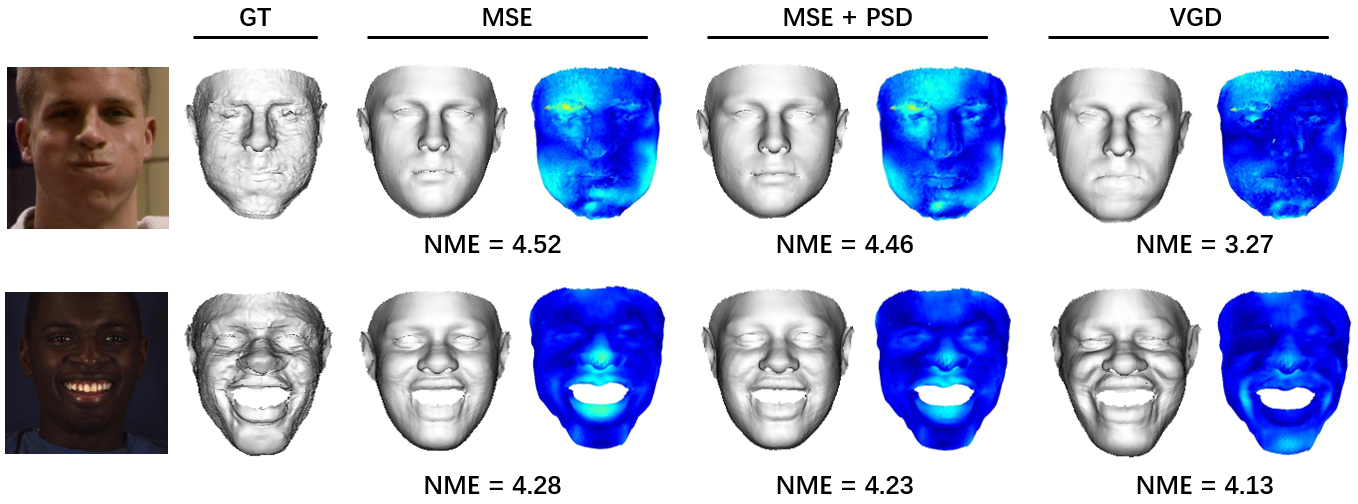}
  \end{center}
  \caption{Visual demonstration with different losses.}
  \label{fig-loss-ablation}
\end{figure}

The results listed in Table~\ref{tab-loss-ablation} indicate that: 1) Even with the ground-truth 3D shape as the supervision, intuitively adopting MSE cannot capture personalized shape well. 2) The introduction of PSD improves the accuracy by directly optimizing the visual effect. However, PSD cannot constrain the face contour due to its defect in back-propagation, which is discussed in Section~\ref{sec-vgd}. 3) The VGD loss further remedies the defects of PSD and captures the face contour well, achieving the best performance. 4) The unsatisfactory results achieved by the combination of PSD and VGD indicate that, it is better to regard PSD as the vertex weights than the loss function. There are also some examples in Fig.~\ref{fig-loss-ablation}.

The PSD draws the 3D face as a plaster sculpture with white vertex color under frontal light, which is common in 3D structure demonstration. We also evaluate the performance of PSD by rendering the 3D face in different vertex colors and lighting conditions, finding little difference.

\begin{table}
\tabcolsep 5pt
	\caption{The Normalized Mean Error (NME) and the Densely Aligned Chamfer Error (DACE) of different losses on the FG3D-test. }
	\begin{center}
		\begin{tabular}{m{1cm}<{\centering}m{1cm}<{\centering}m{1cm}<{\centering}m{1cm}<{\centering}m{1cm}<{\centering}}
			\toprule[1.5pt]
            \multicolumn{3}{c}{Loss Function} & \multirow{2}{*}[-0.5ex]{NME} & \multirow{2}{*}[-0.5ex]{DACE}\\
            \cmidrule(lr){1-3}
		    MSE &  PSD & VGD &  & \\
            \midrule[1pt]
            $\checkmark$   &   &   &  2.80 &  1.50 \\
            $\checkmark$   &  $\checkmark$ &   & 2.75  &  1.47 \\
            &  $\checkmark$ & $\checkmark$   &  2.72 & 1.47 \\
            &   & $\checkmark$  & \textbf{2.66} & \textbf{1.43}  \\
            \bottomrule[1.5pt]
		\end{tabular}
	\end{center}
	\label{tab-loss-ablation}
\end{table}

\subsection{Comparison Experiments}

\begin{figure*}[htb]
\begin{center}
    \includegraphics[width=0.95\textwidth]{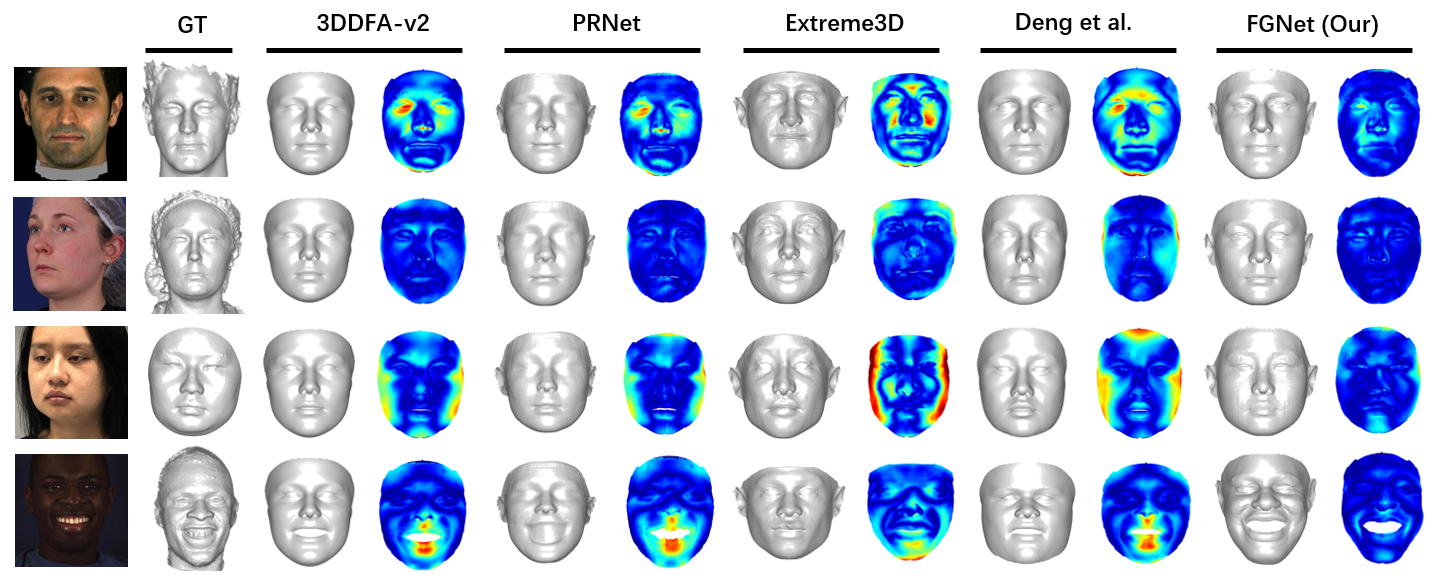}
  \end{center}
  \caption{Qualitative comparison of our method with the representative method of 3DMM fitting (3DDFA-v2), vertex regression (PRNet), shape from shading (Extreme3D) and analysis by synthesis (Deng et.al).}
  \label{fig-quanti-compare}
\end{figure*}

\subsubsection{Quantitative Comparison}
\textbf{Protocol:} To quantitatively compare our method with prior works, we evaluate the single-view 3D reconstruction performance on Florence~\cite{bagdanov2011florence}, NoW~\cite{Sanyal_2019_CVPR} and Stirling/ESRC~\cite{Stirling/ESRC}, where the ground-truth 3D shape is available. For fairness, the FG3D-test is not employed due to its similar capture environment with our training data. We evaluate Florence and NoW according to different yaw intervals, and only report the mean accuracy of Stirling/ESRC due to its limited pose variations. The accuracy is measured by the Densely Aligned Chamfer Error (DACE) due to the difference in mesh topology.

\textbf{Counterparts:} The compared 3D face reconstruction approaches include 3DMM fitting in supervised~\cite{zhu2019face,guo2020towards} and weakly-supervised~\cite{deng2019accurate,yang2020facescape} manners, vertex regression~\cite{16_feng2018joint}, Shape from Shading~\cite{tran2018extreme} and Non-linear 3DMM~\cite{tran2019learning}.

3DDFA~\cite{zhu2019face} is a representative 3DMM fitting method that regresses 3DMM parameters in a supervised fashion, which is recently improved by 3DDFA-v2~\cite{guo2020towards} in generalization through meta-learning. FaceScape~\cite{yang2020facescape} fits a new 3DMM with higher geometric quality than 3DDFA. Deng et al.~\cite{deng2019accurate} further introduce weakly-supervised learning that incorporates low-level and perception-level information.  PRNet~\cite{16_feng2018joint} bypasses 3DMM by directly regressing all the vertex coordinates in one propagation, which potentially helps the network cover a larger shape space than the linear model. Extreme3D~\cite{tran2018extreme} attempts to recover the geometric details by Shape from Shading (SfS). Despite the impressive recovered wrinkles and pores, SfS concentrates on the fine-level details. Their global shapes, such as face contour and facial feature topology, still originate from a fitted 3DMM. Non-linear 3DMM~\cite{tran2019learning} achieves a certain breakthrough by learning a non-linear face model in an analysis-by-synthesis manner, which covers a larger shape space than linear 3DMMs.

\begin{figure*}[!htb]
  \centering
  \subfigure[Florence]{
  \label{fig-ced-compare-florence}
  \includegraphics[width=0.3\textwidth]{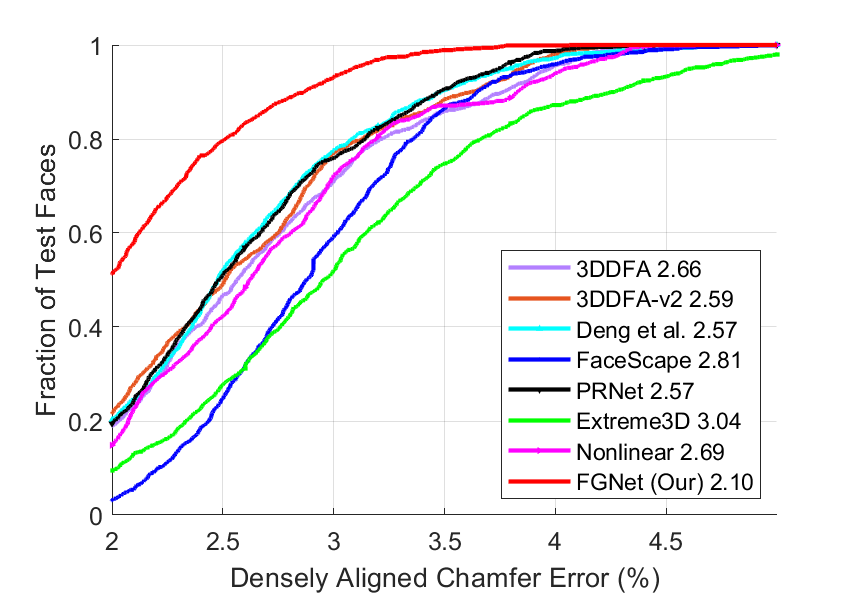}}
  \subfigure[NoW]{
  \label{fig-ced-compare-now}
  \includegraphics[width=0.3\textwidth]{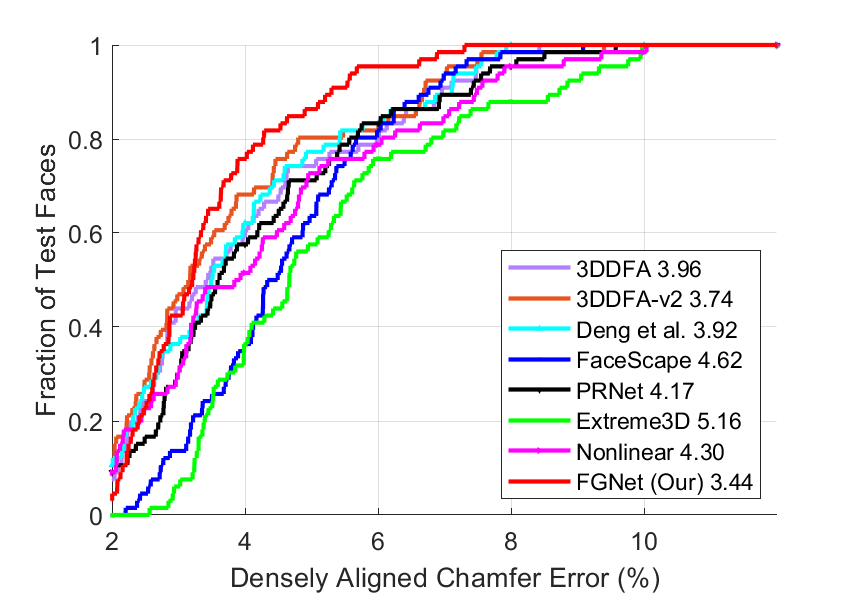}}
  \subfigure[Stirling/ESRC]{
  \label{fig-ced-compare-esrc}
  \includegraphics[width=0.3\textwidth]{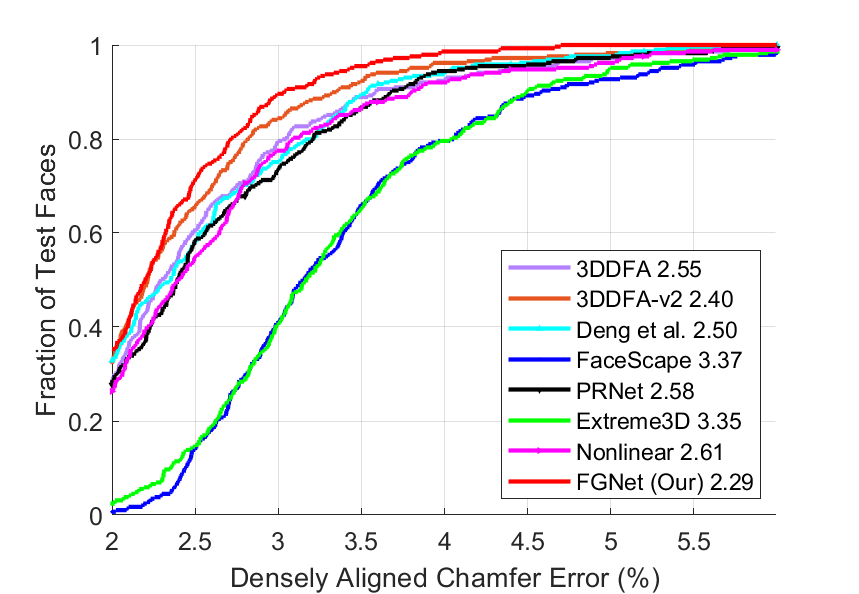}}
  \caption{Comparisons of cumulative error distribution (CED) curves on Florence, NoW and Stirling/ESRC. The error is measured by the mean of DACE.}
  \label{fig-ced-compare}
\end{figure*}

\textbf{Results:} Table~\ref{tab-compare} lists the comparison results and Fig.~\ref{fig-ced-compare} shows the corresponding Cumulative Errors Distribution (CED) curves. Taking Florence as a representative, there are several interesting observations when only shape error is evaluated: 1) Although 3DDFA-v2 is far better than 3DDFA when evaluating the projected 3D faces~\cite{guo2020towards}, their shapes are close. 2) As a non-linear model, PRNet performs similarly to the linear 3DDFA-v2, since its output is still limited by linear 3DMM due to its data-driven characteristics. 3) With the analysis-by-synthesis strategy, Deng's method ranks top among the state-of-the-art methods, demonstrating the benefits of face appearance optimization. 4) We find little difference when comparing Extreme3D with its base model (without details), indicating that the fine-level details acquired by Shape-from-Shading do not change face shapes much. 5) Our method achieves the best result, validating the feasibility of reconstructing personalized shapes in a supervised manner. In addition, we train our model $5$ times and evaluate the performance variance on Florence. Compared with the variance of $0.018$, the improvement of the best baseline ($2.57$ to $2.10$) is more significant, validating the effectiveness of our method.

\begin{table*}
    \centering
    \caption{The Densely Aligned Chamfer Error (DACE) on Florence, NoW and Stirling/ESRC, evaluated by different yaw ranges.}
      \resizebox{0.975\textwidth}{!} {
      \begin{tabular}{cccccccccc}
      \toprule[2pt]
      \multirow{2}{*}[-0.5ex]{Method}  & \multicolumn{4}{c}{Florence} & \multicolumn{4}{c}{NoW} &\multirow{2}{*}[-0.5ex]{Stirling/ESRC}\\
      \cmidrule(lr){2-5} \cmidrule(lr){6-9}      & $[0, 15]$ & $[15, 30]$ & $[30, 45]$ & Mean  & $[0, 15]$ & $[15, 30]$ & $[30, 45]$ & Mean &  \\
      \midrule[1pt]
      3DDFA~\cite{zhu2019face}  &  2.64 & 2.66  & 2.69  & 2.66 &  3.86 &  3.82 &  4.09 & 3.96 & 2.55    \\
      3DDFA-v2~\cite{guo2020towards} & 2.57 & 2.58 & 2.63 & 2.59  & 3.77  &   3.65   & 3.75  &  3.74 & 2.40   \\
      Deng et al.~\cite{deng2019accurate} &  2.48 & 2.59 & 2.67 & 2.57  &  3.66 & 3.57  & 4.22  & 3.92 &  2.50  \\
      FaceScape~\cite{yang2020facescape} & 2.88 & 2.73 & 3.15 & 2.81 & 3.67 & 4.02  & 5.49 & 4.62 & 3.37    \\
      PRNet~\cite{16_feng2018joint}& 2.50 & 2.53 & 2.71 & 2.57 & 4.10 & 3.85 & 4.33 & 4.17 & 2.58 \\
      Extreme3D~\cite{tran2018extreme}&  2.93 & 3.07  & 3.16 & 3.04 & 5.17 & 5.05 & 5.20 & 5.16 & 3.35  \\
      Non-linear 3DMM~\cite{tran2019learning}& 2.74 & 2.66 & 2.66 & 2.69 & 4.73 & 3.96 & 4.13 & 4.30 & 2.61 \\
      \textbf{FGNet (Ours)} & \textbf{2.12} & \textbf{2.05} & \textbf{2.15} & \textbf{2.10} & \textbf{3.45} & \textbf{3.39} & \textbf{3.45} & \textbf{3.44} & \textbf{2.29} \\
      \bottomrule[1.5pt]
      \end{tabular}
      }
    \label{tab-compare}%
\end{table*}%

\begin{figure}[htb]
\begin{center}
    \includegraphics[width=0.48\textwidth]{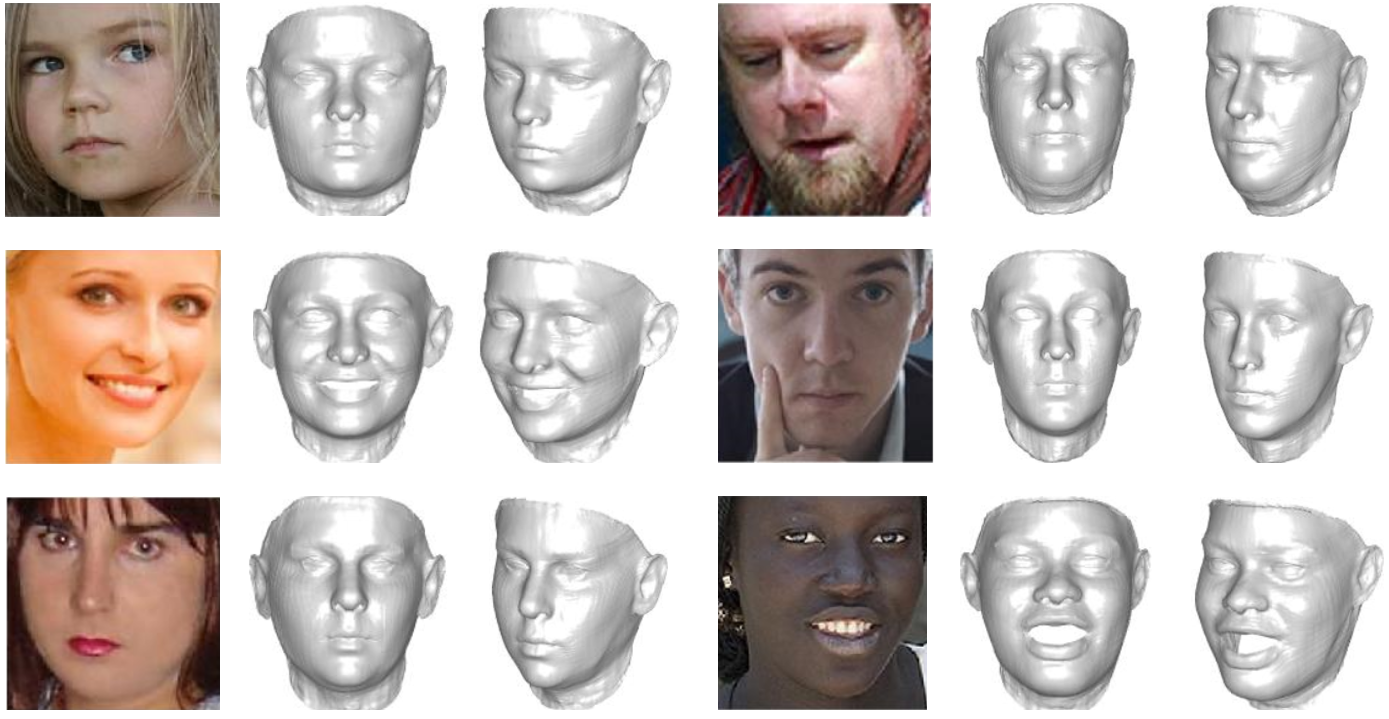}
  \end{center}
  \caption{Qualitative results of our method in unconstrained environment.}
  \label{fig-quali-compare}
\end{figure}

\subsubsection{Qualitative Evaluation}
We present some visual comparisons in Fig.~\ref{fig-quanti-compare} to illustrate the identifiability of the reconstructed shapes. Comparing several representative methods, including 3DMM fitting (3DDFA-v2), vertex regression (PRNet), shape from shading (Extreme3D) and analysis by synthesis (Deng's method), we can first see that the results of 3DDFA-v2 and PRNet are not discriminative enough due to the limitation of linear 3DMM. It is difficult to distinguish the 3D faces by observing the 3D geometry only. Second, Extreme3D recovers plausible geometric details by the shape-from-shading method, but their global shapes do not look identical to the corresponding person. Third, the analysis-by-synthesis method employed by Deng's method improves the visual effect by optimizing the face appearance. Finally, compared with the aforementioned methods, our proposal improves shape accuracy by three advantages: 1) straightforward training signals from the ground-truth 3D shapes, 2) a specific non-linear neural network modeling personalized shape, and 3) a visual-effect-guided loss highlighting shape errors. In the experiments, 3DDFA, 3DDFA-v2, PRNet, Extreme3D, Deng's method and FaceScape are implemented by the released codes, and the non-linear 3DMM is reproduced and trained on 300W-LP~\cite{zhu2019face}.

\begin{figure}[!htbp]
\centering
  \subfigure[Reconstruction across poses]{\label{fig-quali-robust-a}
    \includegraphics[width=0.45\textwidth]{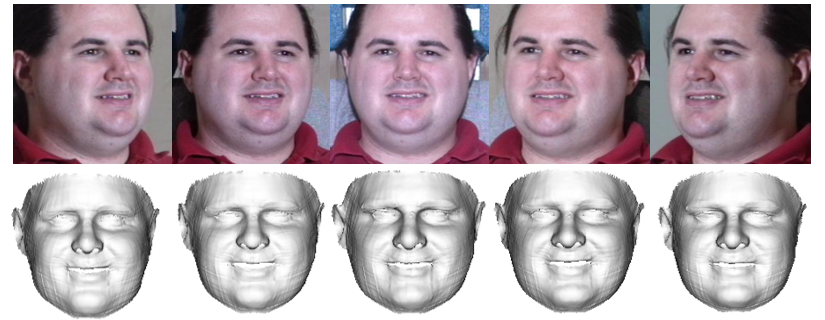}}
  \subfigure[Reconstruction in different illumination.]{\label{fig-quali-robust-b}
    \includegraphics[width=0.45\textwidth]{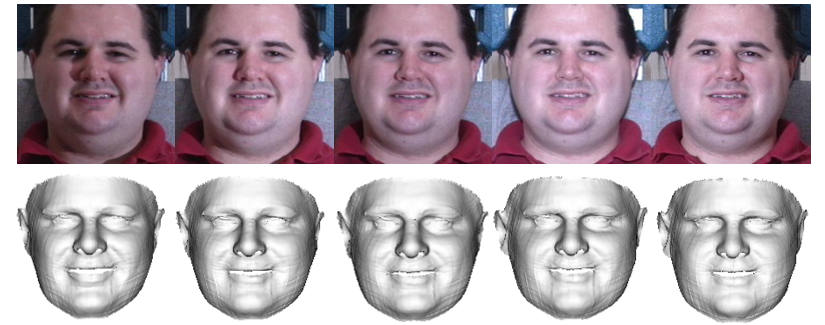}}
  \subfigure[Reconstruction in occlusion]{\label{fig-quali-robust-c}
    \includegraphics[width=0.45\textwidth]{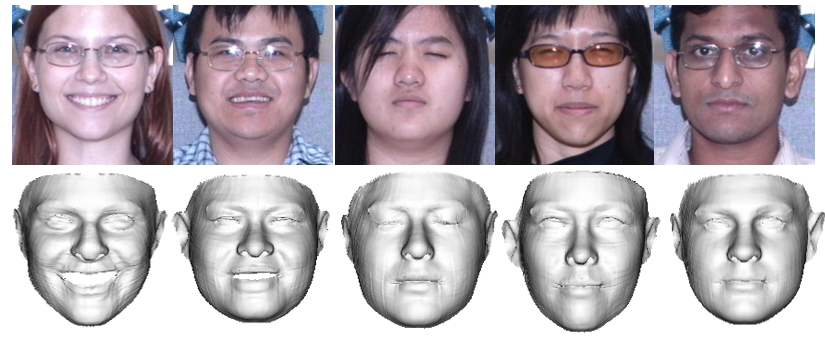}}
  \caption{Qualitative evaluation of the robustness to (a) pose, (b) illumination, and (c) occlusion.}
  \label{fig-quali-robust}
\end{figure}

We further qualitatively evaluate the robustness to pose, illumination and occlusion in Fig.~\ref{fig-quali-robust}. The results show that our method performs well under different poses, side-light and common occlusions such as hair and eye-glasses. The robustness comes from the diverse pose and illumination variations in the training data and the sophisticated ICP registration that filters out occluded vertex matching.

We also attempt to reconstruct the high-fidelity shape in unconstrained environment, evaluated by the samples in AFLW~\cite{kostinger2011annotated}. The results in Fig.~\ref{fig-quali-compare} indicate that, trained on the indoor-collected and 3D-augmented samples, our model generalizes well in an outdoor environment.

\section{Conclusion}

This paper proposes a complete solution for high-fidelity 3D face reconstruction, from data construction to neural network training. With the proposed Virtual Multiview Network (VMN), the input image is rendered at $5$ calibrated views so that pose variations are normalized with little image information lost. Then, the features extracted from multiple views are fused and regressed to a UV displacement map by a novel many-to-one hourglass network. A novel Plaster Sculpture Descriptor (PSD) is also proposed to model the visual effect, which considers the reconstructed shape as a white plaster and measures the similarity between the multiview images rendered from the shape. Besides, to provide abundant samples for network training, we propose to register RGB-D images followed by pose and shape augmentation. Extensive experimental results substantiate the state-of-the-art performance of our proposal on several challenging datasets.

\section*{ACKNOWLEDGMENTS}
This work was supported in part by the National Key Research \& Development Program (No. 2020AAA0140002), Chinese National Natural Science Foundation Projects \#62176256, \#61876178, \#61872367, \#61976229, \#62106264, the Youth Innovation Promotion Association CAS (\#Y2021131) and the InnoHK program.

\small
\bibliographystyle{IEEEtran}
\bibliography{egbib}

\begin{IEEEbiography}[{\includegraphics[width=1in,height=1.25in,clip,keepaspectratio]{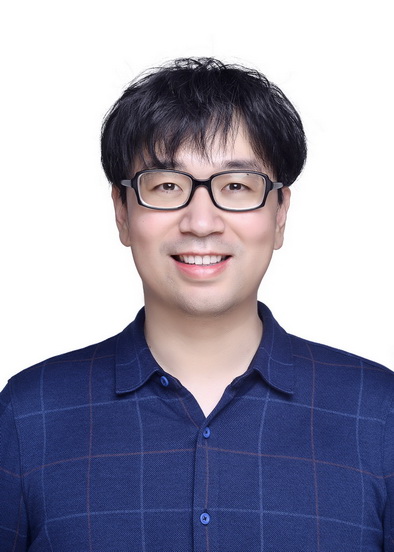}}]{Xiangyu Zhu}
received the BS degree in Sichuan University (SCU) in 2012, and the PhD degree from Institute of Automation, Chinese Academy of Sciences, in 2017, where he is currently an associate professor. His research interests include pattern recognition and computer vision, in particular, image processing, 3D face model, face alignment and face recognition. He is a senior member of the IEEE.
\end{IEEEbiography}
\vspace{-10mm}

\begin{IEEEbiography}[{\includegraphics[width=1in,height=1.25in,clip,keepaspectratio]{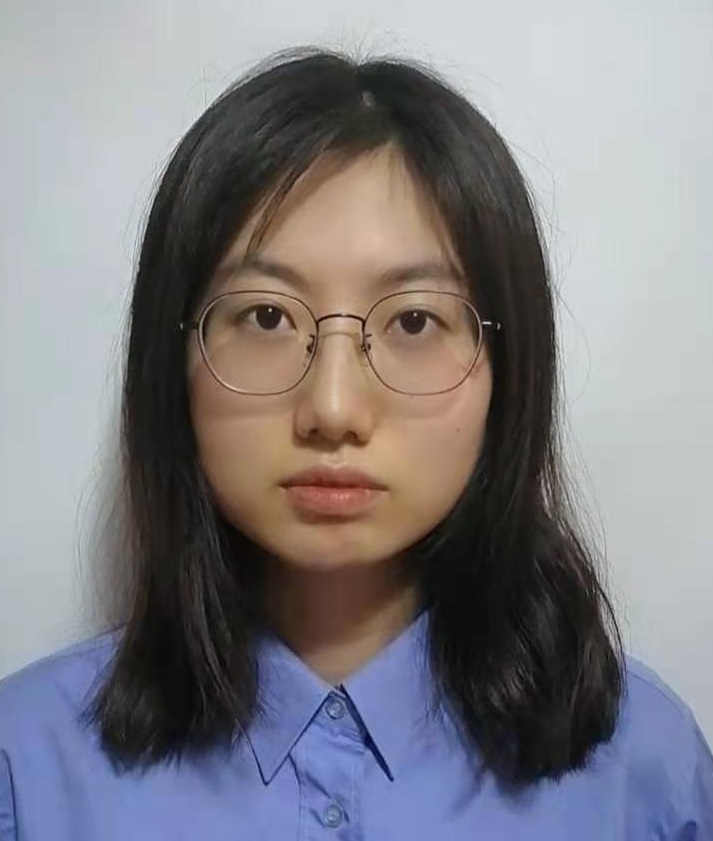}}]{Chang Yu}
	received the BS degree in Xi'an Jiaotong University (XJTU) in 2019. She is working toward the PhD degree in Institute of Automation, Chinese Academy of Sciences. Her research interest includes computer vision, pattern recognition and 3D face reconstruction.
\end{IEEEbiography}
\vspace{-10mm}

\begin{IEEEbiography}[{\includegraphics[width=1in,height=1.25in,clip,keepaspectratio]{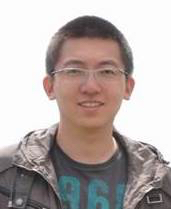}}]{Di Huang}
	received the B.S. and M.S. degrees in computer science from Beihang University, Beijing, China, and the Ph.D. degree in computer science from the \'Ecole centrale de Lyon, Lyon, France, in 2005, 2008, and 2011, respectively. He joined the Laboratory of Intelligent Recognition and Image Processing, School of Computer Science and Engineering, Beihang University, as a Faculty Member. He is currently a Professor with the research interests on biometrics, in particular, on 2D/3D face analysis, image/video processing, and pattern recognition.
\end{IEEEbiography}
\vspace{-10mm}

\begin{IEEEbiography}[{\includegraphics[width=1in,height=1.25in,clip,keepaspectratio]{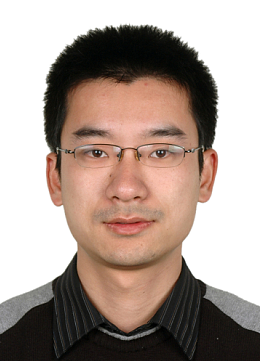}}]{Zhen Lei}
received the BS degree in automation from the University of Science and Technology of China, in 2005, and the PhD degree from the Institute of Automation, Chinese Academy of Sciences, in 2010, where he is currently a professor. He has published more than 200 papers in international journals and conferences. His research interests are in computer vision, pattern recognition, image processing, and face recognition in particular. He is the winner of 2019 IAPR YOUNG BIOMETRICS INVESTIGATOR AWARD. He is a senior member of the IEEE.
\end{IEEEbiography}
\vspace{-10mm}

\begin{IEEEbiography}[{\includegraphics[width=1in,height=1.25in,clip,keepaspectratio]{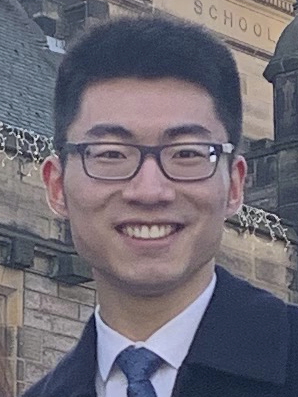}}]{Hao Wang}
received his B.Eng from Beijing University of Posts and Telecommunications, China, in 2018, and MSc from The University of Edinburgh, the UK, in 2019. He is currently a research intern at the Institute of Automation, Chinese Academy of Sciences. His research interests include computer vision, pattern recognition, image processing, human face analysis, biometrics, and 3D shape modeling.
\end{IEEEbiography}
\vspace{-10mm}

\begin{IEEEbiography}[{\includegraphics[width=1in,height=1.25in,clip,keepaspectratio]{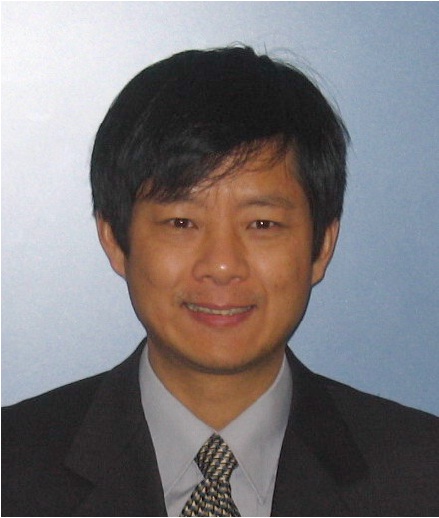}}]{Stan Z. Li}
received the BEng degree from Hunan University, China, the MEng degree from National University of Defense Technology, China, and the PhD degree from Surrey University, United Kingdom. He is currently a chair professor in Westlake University and a guest professor of Center for Biometrics and Security Research (CBSR), Institute of Automation, Chinese Academy of Sciences (CASIA). He was with Microsoft Research Asia as a researcher from 2000 to 2004. Prior to that, he was an associate professor in the Nanyang Technological University, Singapore. His research interests include pattern recognition and machine learning, image and vision processing, face recognition, biometrics, and intelligent video surveillance. He has published more than 300 papers in international journals and conferences, and authored and edited eight books. He was an associate editor of the IEEE Transactions on Pattern Analysis and Machine Intelligence and is acting as the editor-in-chief for the Encyclopedia of Biometrics. He served as a program co-chair for the International Conference on Biometrics 2007, 2009, 2013, 2014, 2015, 2016 and 2018, and has been involved in organizing other international conferences and workshops in the fields of his research interest. He was elevated to IEEE fellow for his contributions to the fields of face recognition, pattern recognition and computer vision and he is a member of the IEEE Computer Society.
\end{IEEEbiography}

\end{document}